\PassOptionsToPackage{table,dvipsnames}{xcolor}
\documentclass{applemlr}

\usepackage{amsmath,amsfonts,bm}

\def\eqref#1{equation~\ref{#1}}

\def\1{\bm{1}}

\DeclareMathAlphabet{\mathsfit}{\encodingdefault}{\sfdefault}{m}{sl}
\SetMathAlphabet{\mathsfit}{bold}{\encodingdefault}{\sfdefault}{bx}{n}

\newcommand{\E}{\mathbb{E}}

\newcommand{\R}{\mathbb{R}}

\usepackage{url}
\usepackage{xurl}

\makeatletter
\AtBeginDocument{\urlstyle{sf}}
\makeatother

\usepackage{tabularx}
\usepackage{booktabs}
\usepackage{listings}
\usepackage{algorithm}
\usepackage{algpseudocode}
\usepackage{subcaption} 
\usepackage{graphicx}
\usepackage{multirow}
\usepackage{amssymb}
\newcommand{\std}[1]{{\scriptsize$\pm$#1}}
\newcommand{\z}{\hphantom{0}}
\usepackage{xspace}
\usepackage{tcolorbox}
\tcbuselibrary{listings,breakable,skins}

\newtcblisting{promptbox}{
  enhanced,
  breakable,
  listing only,
  title={Prompt},
  title after break={Prompt (continued)},
  fonttitle=\small\bfseries,
  coltitle=black,
  colbacktitle=black!7,
  colback=black!2,
  colframe=black!7,
  boxrule=0pt,
  arc=2pt,
  left=8pt, right=8pt,
  top=6pt, bottom=6pt,
  listing options={
    basicstyle=\ttfamily\footnotesize,
    breaklines=true,
    breakatwhitespace=true,
    breakautoindent=false,
    columns=fullflexible,
    keepspaces=true,
    showstringspaces=false,
    numbers=none,
    frame=none
  }
}

\newcommand{\rised}{\textsc{RISED}\xspace}

\title{RISED: RubrIcs for agentic multi-environment Selection and sElf-Distillation}

\author[1,*]{Jingtan Wang}
\author[2]{Sirajul Salekin}
\author[2]{Young mok Jung}
\author[2]{Javier Movellan}
\author[1]{Bryan Kian Hsiang Low}
\author[2]{Manjot Bilkhu}

\affiliation[1]{National University of Singapore}
\affiliation[2]{Apple}

\contribution[*]{Work done during an internship at Apple}

\abstract{
Training a single LLM agent jointly across diverse interactive environments has attracted increasing attention as a route to generalist agents. Existing curriculum and data-selection strategies often allocate training at the environment level or prioritize local reward-based signals, without explicitly considering relationships between current rollouts across environments for prompt-group selection. Meanwhile, as environments are learned at different rates, all-failure and all-success rollout groups can coexist within a batch, leaving those data without group-relative reward signals. Both challenges highlight limitations of relying solely on scalar rewards in multi-environment RL: they provide limited information about cross-environment relationships and no within-group reward contrast when rewards are identical. This motivates richer textual feedback, such as rubrics describing rollout behaviours, to guide learning.
Beyond rubrics' usage as reward, we repurpose rubrics to guide both online data selection and policy supervision. An LLM judge tags each rollout using a predefined rubric vocabulary shared across environments. The resulting profiles guide the selection of data that aligns with the overall behavioural composition of the mixed-environment batch while limiting overlap with already-selected data. Available positive rubrics (describing desired behaviours) provide privileged context for an on-policy self-distillation teacher, supplying additional token-level supervision, while negative rubrics (describing undesired behaviours) guide subsequent rollout generation away from recurring failure modes.
Together, these components form \rised. Across model backbones, \rised achieves the highest mean pass rate across environments and ranks first or second in every individual environment. Rubric-based analysis of \rised can further characterize the behavioural changes accompanying these gains. 

}

\metadata[Correspondence]{\sffamily Jingtan Wang: \url{jingtan.w@u.nus.edu}, Manjot Bilkhu: \url{mbilkhu@apple.com}}
\date{\sffamily\today}

\begin{document}

\maketitle

\section{Introduction}
Training a single LLM agent across diverse interactive environments offers a promising route toward generalist agents~\citep{bercovich2025llama,ren2026magicagent}. Joint training can benefit from transfer across environments and achieve performance comparable to separately trained specialists~\citep{zeng2024agenttuning,zhang2025agentrl}. 
We study joint training on mixed-environment batches, where multiple environments contribute to each policy update~\citep{blakeman2025nemotron, zhang2025agentrl}. Within a limited update budget, an important question is which prompt groups to select from the current rollouts to improve joint training.

Existing approaches address different aspects of training-data allocation. Curriculum methods typically allocate training across domains or environments~\citep{chen2025self, yang2026nemotron, yang2026transferability, zhu2026beyond}, or adjust sample-level priorities using periodically refreshed difficulty estimates~\citep{blakeman2025nemotron}. Online selection instead operates on current rollouts and can adapt to the evolving policy, but commonly prioritizes individual examples using local signals such as reward variability or estimated difficulty~\citep{jiang2025vcrl,sun2025improving, yu2026dapo}. Neither explicitly accounts for relationships between current rollouts from different environments. A second challenge arises from reward-based supervision. Under GRPO, identical rollout rewards provide no within-group reward contrast to distinguish which behaviours should be reinforced. In multi-environment training, all-failure and all-success groups can occur frequently because environments are learned at different rates. In our experiments, they constitute more than half of the candidate prompt groups, leaving substantial training data without a group-relative reward signal.

\begin{figure}[t]\centering
\includegraphics[width=\linewidth]{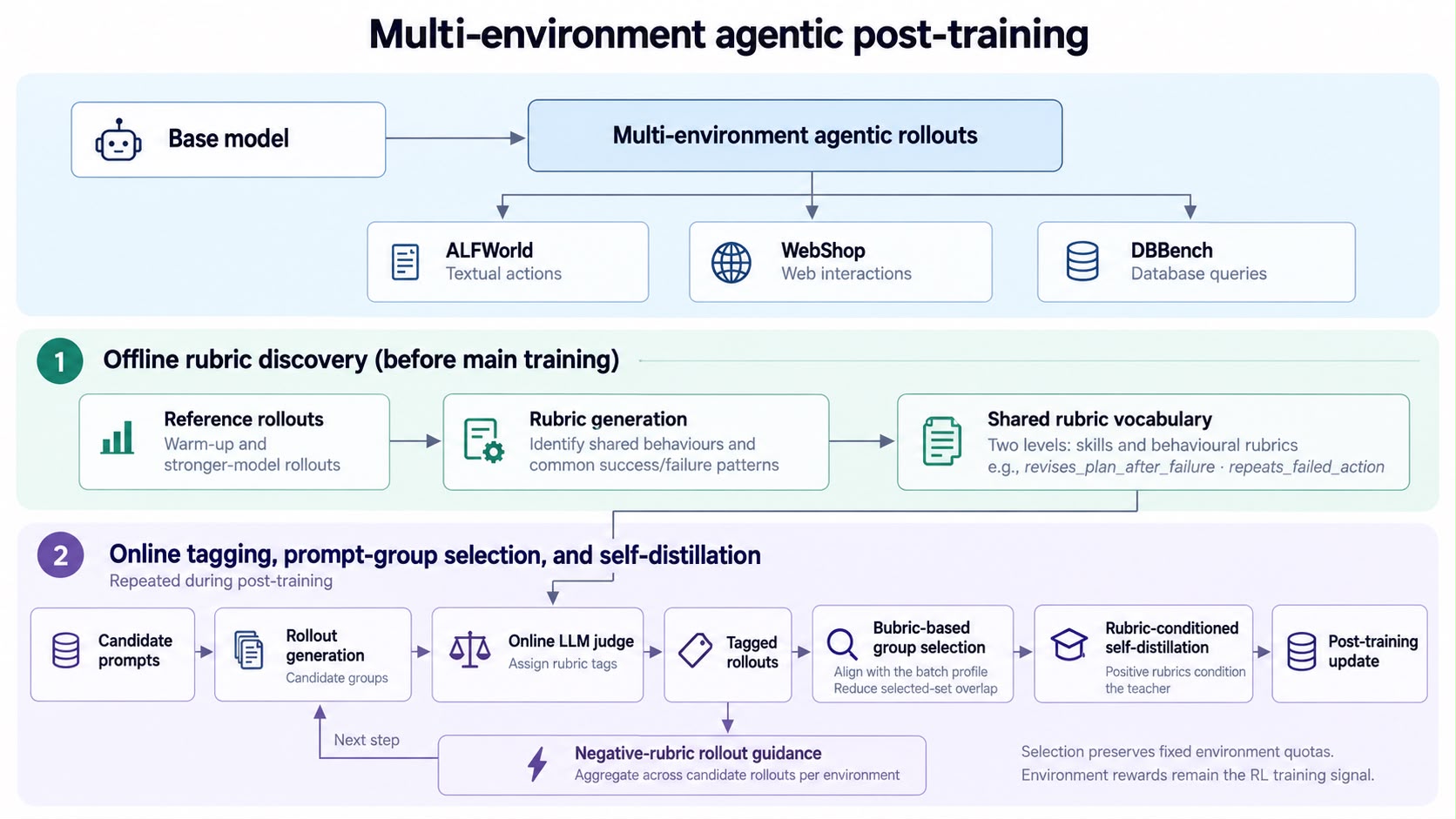}
\caption{Overview of \rised. Sec.~\ref{sec:rubrics:build} describes the offline rubric discovery. Sec.~\ref{sec:rubrics:select} and Sec.~\ref{sec:rubrics:distill} describe the selection and distillation, respectively. Together, these form \rised, which uses rubrics to select training data across diverse agentic environments and reuses them for on-policy self-distillation. }
\label{fig:pipline}
\end{figure}

Both challenges highlight limitations of relying solely on scalar reward signals in multi-environment RL. Reward-level statistics may be too coarse to characterize relationships relevant to cross-environment transfer, while identical rewards provide no within-group reward contrast. Recent work has explored textual feedback, such as rubric-based guidance, as richer supervision than scalar rewards and demonstrated its effectiveness for training LLMs and agents~\citep{gunjal2026rubrics,song2026expanding,zhang2026cm2}. We find that rubric-based feedback is useful beyond policy supervision: it can serve as an effective online data-selection signal for capturing cross-environment relationships, while the same annotations can supply complementary supervision for groups with identical rewards~\citep{bablani2026rubrics, wang2026skill}.

Concretely, an LLM judge tags each rollout using a shared, predefined rubric vocabulary (Sec.~\ref{sec:rubrics:build}), producing rubric profiles that make rollout behaviours comparable across environments.
We also investigate whether rubric profile similarity is informative about cross-environment transfer: our preliminary analysis finds that it correlates more strongly with cross-environment reward improvements than gradient similarity (Sec.~\ref{sec:rubrics:val}). 
Motivated by this finding, we select prompt groups that align with the overall rubric profile of the mixed-environment batch while limiting overlap with already-selected groups (Sec.~\ref{sec:rubrics:select}).
Additionally, the same rubric information is reused during post-training. Available positive rubrics (describing desired behaviours) provide privileged context for an on-policy self-distillation teacher, yielding additional token-level supervision even when rollout rewards are identical, while negative rubrics (describing undesired behaviours) guide subsequent rollout generation away from recurring failure modes (Sec.~\ref{sec:rubrics:distill}).
Together, these components form \rised: \underline{R}ubr\underline{I}cs for agentic multi-environment \underline{S}election and s\underline{E}lf-\underline{D}istillation, with an overview of the framework in Fig.~\ref{fig:pipline}. 
Across model backbones, \rised achieves the highest mean pass rate across environments, while component ablations demonstrate the complementary benefits of selection and self-distillation.
Rubric-based analysis further reveals more frequent positive behaviours than in baseline rollouts, providing an interpretable account of the behavioural changes accompanying these gains.

Our contributions are:
\begin{itemize}
\item We adopt rollout-level rubric profiles as behavioural descriptions (Sec.~\ref{sec:rubrics:build}) and empirically show that rubric similarity correlates more strongly than gradient similarity with cross-environment reward improvements (Sec.~\ref{sec:rubrics:val}).

\item We propose \rised, a framework for multi-environment agentic RL that uses the same predefined rubric vocabulary across environments for online prompt-group selection and self-distillation. The selection component uses rubric profiles to select prompt groups across environments (Sec.~\ref{sec:rubrics:select}). Positive rubrics condition the self-distillation teacher, while negative rubrics guide subsequent rollout generation (Sec.~\ref{sec:rubrics:distill}).

\item We conduct extensive experiments and ablations to evaluate \rised across multiple model backbones (Sec.~\ref{sec:exp:main}) and demonstrate the complementary benefits of selection and self-distillation. We further use rubric profiles to characterize the behavioural changes accompanying these gains (Sec.~\ref{sec:exp:ablation}).
\end{itemize}

\section{Rubrics For Data Selection}
\subsection{Problem Setup}
\label{sec:rubrics:setup}

We train a single policy jointly over a set of environments
\(\mathcal E\) using GRPO~\citep{shao2024deepseekmath}. 
Prompts are scheduled for rollout
generation in a round-robin manner across environments,
targeting a uniform environment mixture, aligned with previous work setting~\citep{zhang2025agentrl}.
At each training step \(t\), we construct a
candidate batch \(\mathcal B_t\) of rollout groups.
Each group \(g=(\tau_{g,i})_{i=1}^{n}\) contains \(n\) rollouts
sharing a prompt \(x_g\) from environment \(e(g)\in\mathcal E\).
A rollout \(\tau_{g,i}=(x_g,y_{g,i})\) consists of the prompt
and a completion or interaction sequence \(y_{g,i}\) generated
by the rollout policy. We oversample by a factor of two, forming \(|\mathcal B_t|=2k\) candidate groups (\(2kn\) rollouts in total), and select a subset \(\mathcal S_t\subseteq\mathcal B_t\) of \(|\mathcal S_t|=k\) groups (\(kn\) rollouts) for the policy update.

Two properties of this setup are worth emphasizing. First, the choice is made \emph{online}, once per step, against the current policy, unlike one-time selection of a fixed training subset, which cannot respond to a policy that keeps changing. Second, selecting from a $2\times$ candidate pool rather than the full task set bounds the cost: in RL, scoring a candidate usually requires generation, so evaluating the entire pool at every step is not affordable. Generating a surplus and selecting down is a standard compromise in online batch selection \citep{loshchilov2015online,mindermann2022prioritized,zou2026utilitydiversity}, and similar patterns are seen in RL post-training \citep{yu2026dapo,zheng2026act}.

\subsection{Rubric Construction}
\label{sec:rubrics:build}

\textbf{From rubrics as rewards to rubrics as descriptions.}
Rubrics decompose rollout quality into individually identifiable behaviours and have primarily been used as reward signals for RL~\citep{huang2026rubicap,jia2026open,rezaei2026online,zhang2026cm2}. 
We repurpose these behavioural descriptions for sample selection, using shared rubric profiles to compare rollouts across environments.
Using them as rewards requires careful calibration: every criterion is eventually folded into a scalar \citep{tyagi2026not}, so a missing or miscalibrated criterion
can provide no gradient toward an omitted behaviour, or the wrong one. This requirement is expensive in agentic environments, where a task may admit many successful rollouts and exhaustively specifying in advance the desired good behaviours is difficult. We ask considerably less of the criteria: we identify the prominent behaviours exhibited by each rollout and compare rollouts through these behavioural profiles.\footnote{For selection, an incomplete rubric set may make rollouts appear more similar, weakening the signal rather than directly defining an incorrect reward. A more important requirement when using rubrics for selection is consistency across rollouts. The criteria need not exhaustively cover desirable behaviours, but should capture enough variation to distinguish rollouts.}

\textbf{Rubric Vocabulary.}
Each rubric denotes a fine-grained behavioural pattern, represented by a \texttt{snake\_case} name. We construct a predefined rubric vocabulary from warm-up model rollouts and augment it with behaviours identified in stronger LLM-generated rollouts.
Because warm-up rollouts predominantly end in failure, successful LLM-generated rollouts are introduced to improve coverage of desired behaviours that would otherwise be underrepresented. Together, these sources broaden coverage of desired and undesired behaviours, described by positive and negative rubrics, respectively. We use \texttt{Gemini-2.5-Pro} to propose rubric names from warm-up rollout behaviours, consolidate semantically similar names, and organize them into skills. We then augment the vocabulary by extracting additional behaviours from teacher rollouts generated by \texttt{Claude-Opus-4.6} and warm-up rollouts again, merging redundant names before incorporating new rubrics. The final vocabulary contains 82 rubrics across 11 skills\footnote{The vocabulary is constructed from warm-up and teacher rollouts across four environments. The main experiments use three of these environments; removing the rubrics and skill specific to the excluded environment yields 59 rubrics grouped into 10 skills.}, covering both environment-general and environment-specific behaviours. Each rubric includes a short description for teacher conditioning and training-time rollout guidance (Sec.~\ref{sec:rubrics:distill}).

\textbf{Rubric Extraction. }
Given a rollout and the two-level vocabulary, an LLM judge identifies a set of $(\text{skill},\text{rubric},\text{role},\text{count})$ tuples from the predefined vocabulary. This gives rollouts from different environments a common behavioural representation while allowing the judge to navigate the vocabulary hierarchically rather than through a single flat list. 
The \texttt{role} indicates whether the behaviour is desirable (\emph{demonstrate}) or undesirable (\emph{violate}), while \texttt{count} records the number of occurrences of that behaviour identified in the rollout. Rubric extraction requires one judge inference call per rollout, without gradient computation. We serve \texttt{Qwen3-32B} locally and tag completed rollouts asynchronously, overlapping rubric annotation with rollout generation. App.~\ref{app:time_complexity} reports the tagging overhead. The complete construction and extraction pipeline, including prompts and example rubric names, is provided in App.~\ref{app:vocab_construct}.

\subsection{Does Rubric Similarity Track Cross-Environment Transfer?}
\label{sec:rubrics:val}

\textbf{Setup.}
We test whether similarity between rubric profiles is associated with cross-environment transfer. We train a policy on one source environment and evaluate it on both the source and held-out environments. We measure reward improvements in each environment to assess the benefits of source-environment training, with improvements on held-out environments indicating cross-environment transfer. A useful similarity signal should assign higher similarity to environments that exhibit larger reward improvements.
We present results with WebShop~\citep{yao2022WebShop} as the source environment in the main text and repeat the analysis with DBBench~\citep{liu2024agentbench} as the source in App.~\ref{app:correlation}. For the WebShop setting, we evaluate on WebShop, DBBench, ALFWorld~\citep{shridhar2021ALFWorld}, and AppWorld~\citep{trivedi2024AppWorld}, covering shopping, database interaction, household object manipulation, and application-based tasks, respectively. Detailed environment descriptions are provided in App.~\ref{app:training_data}.

Every $25$ training steps up to step $500$, we freeze the policy and collect $512$ fresh WebShop training rollouts ($64$ prompts $\times$ $8$ samples), together with four rollouts per validation prompt from each environment. For this analysis, we use a fixed set of up to 200 validation prompts per environment across all checkpoints, retaining the full set when fewer than 200 are available. Each rollout is annotated with its corresponding rubrics. For every checkpoint--environment pair $(t,e)$, we measure the reward improvement
$
\Delta_t(e)=\bar r_t(e)-\bar b_e,
$
where \(\bar r_t(e)\) is the mean validation reward in
environment $e$ at checkpoint \(t\), and \(\bar b_e\) is the mean reward at the first measured checkpoint. We exclude the first checkpoint because $\Delta_{t_0}(e)=0$ by construction. 
We compare this reward-improvement signal against two alignment signals: the cosine similarity between training and validation rubric profiles and, as a baseline, the cosine similarity between the corresponding training and validation gradients.

\textbf{Rubric-based signals.}
Let $\mathcal R=\{r_1,\ldots,r_m\}$ denote the rubric vocabulary.  
For each rollout $\tau$,
let $c_r(\tau)$ denote the judge-assigned count of rubric $r$,
with zero assigned when $r$ is absent.
For any collection of rollouts \(\mathcal A\), define its \emph{rubric profile} as the count vector: $
\mathbf q(\mathcal A)
=
\left(
\sum_{\tau\in\mathcal A}c_{r_1}(\tau),
\ldots,
\sum_{\tau\in\mathcal A}c_{r_m}(\tau)
\right)^{\!\top}
$.
Let \(\mathcal T_t\) denote the WebShop training rollouts at step \(t\) and \(\mathcal V_t(e)\) the validation rollouts from environment \(e\). We define rubric similarity as
$
s_t^{\mathrm{rubric}}(e)
=
\mathrm{sim}\!\left(
\mathbf q(\mathcal T_t),
\mathbf q(\mathcal V_t(e))
\right),
$ where $\mathrm{sim}$ denotes cosine similarity, defined as
zero when either profile is zero. 
Because cosine similarity is invariant to positive scaling, this signal captures the relative composition of tagged behaviours rather than the number of aggregated rollouts. We additionally compute positive-only and negative-only
profiles using the same rubric coordinates, including
$c_r(\tau)$ only when the corresponding role is
\textit{demonstrate} or \textit{violate}, respectively,
and assigning zero otherwise. Skill-level profiles aggregate counts over rubrics belonging
to the same skill.
Tab.~\ref{tab:rho-3env} reports Spearman and Pearson correlations between
each signal and reward improvement, computed across
checkpoint--environment pairs. We report results both across all four environments and over the three held-out environments, excluding WebShop.

  \begin{table}[t]\centering\small
  \caption{Correlation between similarity signals and reward
improvement under WebShop-only training. Spearman ($\rho$)
and Pearson ($r$) correlations are pooled over
checkpoint--environment pairs across steps. Held-out results exclude WebShop.
Positive-only, negative-only, and combined profiles are
denoted by pos, neg, and all, respectively.}
  \label{tab:rho-3env}
  \begin{tabular}{llccccccc}
  \toprule
  & & \multicolumn{3}{c}{Rubric-level} & \multicolumn{3}{c}{Skill-level} & Gradient \\
  \cmidrule(lr){3-5}\cmidrule(lr){6-8}\cmidrule(lr){9-9}
  Val.\ envs & & pos & neg & all & pos & neg & all & full layer. \\
  \midrule
  Held-out           & $\rho$ & $-0.52$ & $+0.04$ & $\mathbf{+0.60}$ & $-0.69$ & $+0.30$ & $+0.43$ & $+0.18$ \\
  & $r$    & $-0.49$ & $+0.07$ & $\mathbf{+0.77}$          & $-0.61$ & $+0.43$ & $+0.52$ & $+0.21$ \\
  \addlinespace
  All env            & $\rho$ & $+0.32$ & $+0.57$ & $\mathbf{+0.81}$ & $+0.25$ & $+0.69$ & $+0.75$ & $+0.28$ \\
  & $r$    & $+0.76$ & $+0.84$ & $\mathbf{+0.87}$          & $+0.69$ & $+0.87$ & $+0.87$ & $+0.27$ \\
  \bottomrule
  \end{tabular}
  \end{table}

\textbf{Results.}
A stronger correlation indicates that a similarity signal more closely tracks which environments improve during WebShop training. 
As shown in Tab.~\ref{tab:rho-3env}, 
the strongest positive correlations are obtained using the combined positive-and-negative rubric profile.
Using both roles provides stronger positive associations, particularly on held-out environments, suggesting that similarities in both successful behaviours and failure modes are informative for cross-environment transfer analysis. Aggregating rubrics into broader skills generally weakens this signal. For example, WebShop and DBBench both contain rubrics belonging to the skill \texttt{tool\_usage\_and\_formatting}. Skill-level aggregation therefore makes them appear similar even when the underlying fine-grained behaviours differ.
Rubric similarity also consistently identifies which environment benefits most from training across all four environments: it identifies the most-improved environment at all $19$ evaluated checkpoints, compared with $8$ checkpoints for gradient similarity. Gradient signal is less strongly correlated with reward improvements in this analysis. 
Additionally, we evaluate the stability of the tagging procedure by independently re-tagging the same rollouts. 
Independent re-tagging yields similar environment-level transfer correlations, with the reported Spearman coefficients changing by at most $0.060$ (App.~\ref{app:reliability}).

\subsection{Rubrics for Online Data Selection}
\label{sec:rubrics:select}

The environment-level analysis motivates using fine-grained rubric counts from both positive and negative rubrics as a shared representation. A fixed vocabulary shared across environments makes these profiles directly comparable across environments and training steps. We now apply this representation to online prompt-group selection. 

\textbf{Selection.}
For each candidate prompt group \(g\in\mathcal B_t\), we aggregate the rubric counts of its \(n\) rollouts into a group profile $\mathbf q(g)
=
\left(
\sum_{i=1}^{n}c_{r_1}(\tau_{g,i}),
\ldots,
\sum_{i=1}^{n}c_{r_m}(\tau_{g,i})
\right)^{\!\top}.$ For the candidate pool, we use the shorthand
$
\mathbf q(\mathcal B_t)
=
\sum_{g\in\mathcal B_t}\mathbf q(g),
$
which summarizes all candidate rollouts at step $t$. Starting from $\mathcal S=\varnothing$, we select groups
greedily in a round-robin manner across environments.
On each environment’s turn, we choose the highest-scoring unselected candidate using an MMR-inspired relevance–redundancy score~\citep{carbonell1998mmr}: 
\begin{equation}
\operatorname{score}(g\mid\mathcal S)
=
\underbrace{
\mathrm{sim}\!\left(
\mathbf q(g), \mathbf q(\mathcal B_t)
\right)
}_{\text{relevance}}
-
\underbrace{
\mathrm{sim}\!\left(
\mathbf q(g),
\textstyle\sum_{g'\in\mathcal S}\mathbf q(g')
\right)
}_{\text{redundancy}}.
\label{eq:rubric-selection}
\end{equation}
We add the chosen group to $\mathcal S$ and continue
until $|\mathcal S|=k$, yielding $\mathcal S_t=\mathcal S$.
The relevance term favours groups aligned with the overall rubric profile of the mixed-environment batch, while the redundancy term penalizes overlap with the selected aggregate. Under round-robin selection, choosing a group in one environment changes the redundancy scores for subsequent choices in others, potentially favouring complementary behaviours as selection process proceeds, such as environment-specific ones. Fixed quotas preserve the baseline environment mixture.
A similar form is common in online batch selection for supervised fine-tuning~\citep{zou2026utilitydiversity} and RL~\citep{lin2026activedpo}, though typically over embeddings rather than rubrics.

\textbf{Rubric selection and reward variance capture different properties.}
Rubric selection does not implicitly act as a reward-variance filter. Over a \(1000\)-step run on three environments, selected groups have mean within-group reward variance \(0.0557\), compared with \(0.0507\) in the candidate pool (\(1.10\times\)), while a variance-targeted selector reaches approximately \(2\times\) in this run. The two signals therefore capture different properties: reward variance indicates whether a group offers within-group reward contrast, whereas rubric overlap describes which behaviours it exercises.

\textbf{Auxiliary supervision for groups with identical rewards.}
Across a $1000$-step run on three environments, $62.9\%$ of candidate groups have zero within-group reward variance, with a similar fraction among rubric-selected groups. Rubric-based selection
therefore retains many groups whose identical rewards provide
no within-group contrast to distinguish which behaviours
should be reinforced.  We then reuse the same rubric annotations for self-distillation to provide complementary token-level supervision on the selected rollouts.

\section{Rubrics For Distillation}
\label{sec:rubrics:distill}

\paragraph{On-policy self-distillation.}
Distillation trains a student to match a teacher's output distribution rather than a hard label, providing a dense per-token target in place of a scalar reward. \emph{On-policy} distillation applies this target to the student's own rollouts, correcting the policy on states it actually visits~\citep{agarwal2024policy}. In \emph{self}-distillation, teacher and student generally share parameters but differ in conditioning: the teacher is the same policy given privileged context $c$ unavailable to the student~\citep{hubotter2026reinforcement}. Thus, no separate teacher is trained, and supervision requires only an additional forward pass. 
For a completion $y$ generated from prompt $z$, self-distillation
conditions the teacher on an additional privileged context $c$ while
leaving the student's input unchanged. Following~\citet{wang2026skill},
we use
\begin{equation}
\label{eq:sdl}
\mathcal L_{\mathrm{SD}}
=
\mathbb E[\rho_j k_3(\ell_j)],
\qquad
\ell_j
=
\log\pi_\theta(y_j\mid z,y_{<j})
-
\log\pi_{\bar\theta}(y_j\mid z\oplus c,y_{<j}),
\qquad
k_3(\ell)=e^{-\ell}-1+\ell,
\end{equation}
where $\bar{\theta}$ denotes the parameter values used for
the teacher forward pass, whose log-probabilities are treated
as constants when differentiating $\mathcal L_{\mathrm{SD}}$.
$\rho_j
=
\frac{
\pi_\theta(y_j\mid z,y_{<j})
}{
\operatorname{sg}[
\pi_{\theta_{\mathrm{old}}}(y_j\mid z,y_{<j})
]
}$
is the importance ratio between the current student and the rollout policy, where $sg$ denotes stop gradient. 
Crucially, this loss provides dense token-level supervision independently of the GRPO-based advantage coefficients.
The overall policy update minimizes
$\mathcal L_{\mathrm{GRPO}}+\lambda\mathcal L_{\mathrm{SD}}$,
where $\mathcal L_{\mathrm{GRPO}}$ follows the AgentRL training
objective~\citep{zhang2025agentrl} and $\lambda$ is the self-distillation coefficient. Both losses are evaluated on the selected rollout groups $\mathcal S$, with token-level terms restricted to generated tokens. Expansion of $\mathcal L_{\mathrm{GRPO}}$ and hyperparameters
are given in App.~\ref{app:train_loss}.

\textbf{Positive rubrics distill; negative rubrics steer.}
The rubrics already tagged for batch selection also provide the privileged context $c$, requiring no additional effort. For each rollout, we take the deduplicated union of its top-\(3\) ($3$ highest-count rubrics) positive rubrics and the top-\(3\) positive rubrics of its group, and render their descriptions as a privileged context for the teacher. Groups without positive rubrics may receive no rubric-derived teacher context (App.~\ref{app:faq:no_positive}). Distillation then trains the student to reproduce the desired behaviour without the privileged context $c$.
The choice to include only positive rubrics in the distillation context is motivated by prior work and supported by our ablation. Prior work similarly constructs rubric-conditioned teachers from the best rollouts~\citep{xia2026enhancing} or compress 
rollouts into separate contextual lessons \citep{xia2026skillrl}. Mixing positive and negative material is not consistently beneficial, with its effect depending on the difficulty and comprehensibility of the context \citep{armandpour2026unmasking}. Our ablation in App.~\ref{app:distill-role} likewise shows that filling context slots with both positive and negative rubrics degrades overall performance. 

Nevertheless, negative rubrics contain valuable information, particularly early in training. On ALFWorld, for example, negative-tagged rollouts outnumber positive-tagged rollouts by more than $10\times$ at step $10$, 
but fall below parity around step $330$. We use negative rubrics to steer the next step's rollouts rather than to condition the teacher. At the end of each step, we aggregate tags over all candidate rollouts from each environment and compare the numbers containing at least one negative rubric and at least one positive rubric. If violations are more prevalent, we prepend that environment's top-$3$ negative rubrics to the next step that environment’s prompts as an ``anti-patterns'' block. 
When this steering is active, $z$ in Eq.~\ref{eq:sdl} denotes the resulting steered prompt. The same $z$ is provided to both student and teacher, while only the positive-rubric context $c$ is privileged to the teacher.
Overall, negative rubrics steer how rollouts are sampled, whereas positive rubrics provide the self-distillation target.
Negative-rubric injection is used only during training. Validation rollouts use the original task prompts, without injected rubrics or privileged teacher context.
Combining selection in Sec.~\ref{sec:rubrics:select} and the distillation here, we propose our approach \rised: \underline{R}ubr\underline{I}cs for agentic multi-environment \underline{S}election and s\underline{E}lf-\underline{D}istillation, as the full algorithm summarized in Alg.~\ref{alg:step} in Appendix. 

\section{Experiments and Results}
\subsection{Experimental Setting}
\label{sec:exp:setup}
We build on AgentRL \citep{zhang2025agentrl}, which provides
a unified function-call interface for multi-turn interaction. We train jointly on three AgentBench environments \citep{liu2024agentbench}: ALFWorld, WebShop, and DBBench.
The three differ in action space, reward mechanics, and rollout length,
which is what makes their interaction non-trivial. An episode is capped at $20$ interaction turns and
$8192$ total tokens, and an unfinished episode receives $-0.2$.
We use GRPO with a batch of $64$ prompts and $n=8$ rollouts per prompt. At each step, the rollout buffer generates $128$ prompt groups, from which we select $64$ (Sec.~\ref{sec:rubrics:select}).
Across all runs, we follow the
AgentRL training protocol, including environment-wise advantage normalization
and stale-policy rollout generation.

We evaluate validation pass rate every $10$ training steps on fixed
held-out sets, sampling $4$ rollouts for each validation prompt.
For each environment, we report the mean pass rate (binary success indicator) across prompts and rollouts. Overall performance is the unweighted mean of the three environment-level pass rates.
To evaluate effectiveness across model backbones, we train \texttt{Qwen2.5-3B-Instruct} (use \texttt{Qwen2.5-3B} later for simplicity) on four nodes with eight H100 GPUs each 
and \texttt{Qwen3-4B} without thinking on eight B200 GPUs. 
Methods requiring rubric annotations, including \rised, its component ablations, and distillation baselines, additionally use a dedicated eight-GPU server for \texttt{Qwen3-32B} judge inference. 
We report main results as the mean and standard deviation over multiple runs. Additional ablation experiments use a single seed unless otherwise stated. Full optimization and rollout
hyperparameters are given in App.~\ref{app:hyper}.

\subsection{Baselines}
\label{sec:exp:baselines}
We compare against five baselines.
\textbf{GRPO-64} uniformly samples $64$ prompts and generates $8$ rollouts per prompt, using all $512$ rollouts per step without selection.
\textbf{GRPO-128} doubles the prompt batch size to $128$ and updates on all $1024$ rollouts.
Like \rised, \textbf{Global Variance} and \textbf{Round-Robin Variance} generate $128$ candidate groups and select $64$ for training based on within-group reward variance.
Reward variance provides an inexpensive proxy for a GRPO learning signal: groups with identical rewards have zero variance. For binary rewards, variance induces the same ordering as mean absolute group-standardized advantage. This motivates our variance-based selectors and relates them to DAPO’s filtering of zero-variance groups~\citep{yu2026dapo}.
Global Variance selects the highest-variance groups across the candidate pool, whereas Round-Robin Variance ranks groups within each environment and interleaves selection across environments, matching \rised's environment scheduling.
Finally, as a distillation-oriented baseline, \textbf{All-Rubric SD} adapts \citet{bablani2026rubrics} to our agentic RL setting by retaining GRPO and conditioning the teacher on all rubrics. Because our environments do not provide predefined task rubrics, these annotations come from our fixed vocabulary. It uses 64 prompt groups per update, without selection or negative-rubric guidance.

\begin{figure}[t]\centering
  \includegraphics[width=\linewidth]{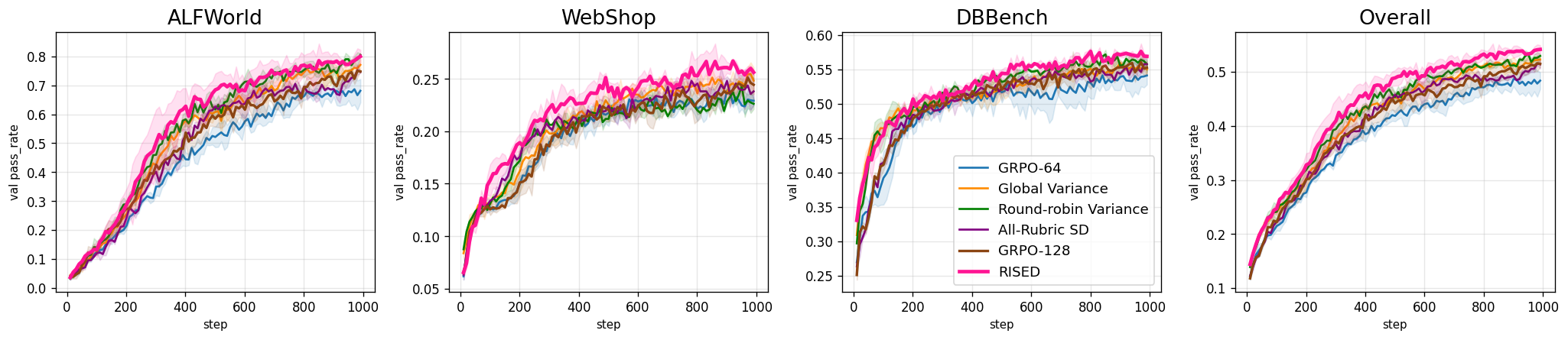}
\caption{\textbf{Validation dynamics for \rised compared with baseline}. Validation pass rate during joint training on ALFWorld, WebShop, and DBBench with \texttt{Qwen2.5-3B-Instruct}, shown per environment and overall. \rised achieves higher validation pass rates than the baselines,
both overall and in each environment, during most of the middle
and later stages of training. }
\label{fig:main:3b}
\end{figure}

\begin{table}
\caption{\textbf{Main results of \rised compared with baselines.} Final-checkpoint pass rates on ALFWorld, WebShop, and DBBench, and overall, for \rised and five baselines on two backbones. \textbf{Bold} and \underline{underlined} values indicate the best and second-best results, respectively, within each backbone. \rised achieves the highest mean overall pass rate and ranks among the top two methods in every individual environment.}
  \begin{tabular}{llccc>{\columncolor{gray!10}}c}
  \toprule
  Model & Method & ALFWorld & WebShop & DBBench & \textbf{Overall} \\
  \midrule
  \multirow{6}{*}{Qwen2.5-3B} & GRPO-64 & 0.681\std{0.041} & 0.229\std{0.004} & 0.541\std{0.013} & 0.484\std{0.015} \\
  & GRPO-128 & 0.748\std{0.000} & 0.244\std{0.009} & 0.552\std{0.009} & 0.515\std{0.006} \\
  & Global Variance & 0.771\std{0.020} & \underline{0.248\std{0.013}} & 0.550\std{0.008} & 0.523\std{0.006} \\
  & Round-robin Variance & \textbf{0.805\std{0.005}} & 0.226\std{0.005} & \underline{0.557\std{0.011}} & \underline{0.530\std{0.003}} \\
  & All-Rubric SD & 0.747\std{0.003} & 0.238\std{0.013} & \underline{0.557\std{0.007}} & 0.514\std{0.008} \\
  & \textbf{\rised} & \underline{0.800\std{0.023}} & \textbf{0.256\std{0.003}} & \textbf{0.569\std{0.002}} & \textbf{0.542\std{0.008}} \\
  \midrule
  \multirow{6}{*}{Qwen3-4B} & GRPO-64 & 0.706\std{0.115} & 0.279\std{0.014} & 0.563\std{0.011} & 0.516\std{0.046} \\
   & GRPO-128 & 0.862\std{0.009} & 0.290\std{0.000} & 0.583\std{0.018} & \underline{0.579\std{0.009}} \\
  & Global Variance & 0.821\std{0.069} & \underline{0.294\std{0.008}} & 0.567\std{0.020} & 0.561\std{0.019} \\
  & Round-robin Variance & \textbf{0.881\std{0.032}} & 0.279\std{0.003} & 0.557\std{0.053} & 0.573\std{0.030} \\
  & All-Rubric SD & 0.841\std{0.008} & 0.284\std{0.005} & \textbf{0.599\std{0.013}} & 0.574\std{0.009} \\
  & \textbf{\rised} & \underline{0.865\std{0.011}} & \textbf{0.306\std{0.012}} & \underline{0.584\std{0.003}} & \textbf{0.585\std{0.001}} \\
  \bottomrule
  \end{tabular}
  \label{tab:main_results}
  \end{table}
  
\subsection{Main Results}
\label{sec:exp:main}

Tab.~\ref{tab:main_results} reports final-checkpoint performance on both backbones, while Fig.~\ref{fig:main:3b} shows the full validation dynamics for \texttt{Qwen2.5-3B}.
\textbf{\rised improves overall performance on both backbones.}
\rised achieves the highest mean overall pass rate on both \texttt{Qwen2.5-3B} and \texttt{Qwen3-4B}, improving over GRPO-64 by $5.8$ and $6.9$ percentage points, respectively. 
Relative to the strongest baseline on each backbone, the gains are $1.2$ and $0.6$ points.
Across individual environments, \rised ranks first or tied for second in all six backbone--environment comparisons. Fig.~\ref{fig:main:3b} further shows that, on \texttt{Qwen2.5-3B}, \rised generally outperforms the baselines both overall and in each environment throughout the middle and later stages of training. Its advantage thus extends beyond just the final checkpoint.
\textbf{\rised reaches baseline performance in fewer optimizer steps.}
As shown in Fig.~\ref{fig:main:3b}, \rised reaches the final overall pass rate of GRPO-64 in approximately half as many optimizer steps. It also reaches the final performance levels of the other baselines earlier in training, although the reductions are smaller.
\textbf{Selecting training data improves on using all candidates.}
\rised outperforms GRPO-128 overall on both backbones, despite updating on only $512$ of the $1024$ generated rollouts per update. These results suggest that rubric-based selection together with distillation can be more effective than increasing the GRPO update batch alone.

\section{Ablations and Further Analysis}
\label{sec:exp:ablation}

\begin{figure}[t]\centering
\includegraphics[width=\linewidth]{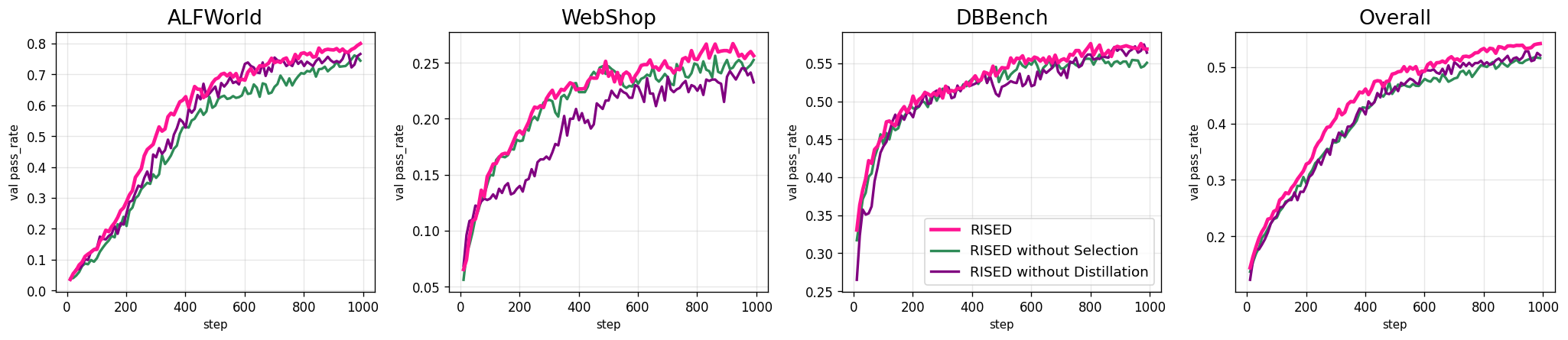}
\caption{\textbf{Component ablations.} Validation pass rate across training steps for \rised (pink), \rised without selection (green), \rised without distillation (purple). Combining selection and distillation yields the strongest overall performance, while their individual contributions vary across environments.}
\label{fig:component_ablation}

\label{fig:ablation}
\end{figure}
\textbf{Component ablations. }
Using \texttt{Qwen2.5-3B}, we separately remove rubric-guided selection (remove Sec.~\ref{sec:rubrics:select} component) and rubric-conditioned distillation (remove Sec.~\ref{sec:rubrics:distill} component means also removing negative steering) from \rised (Fig.~\ref{fig:component_ablation}).
Removing either component reduces overall performance, supporting their complementary roles: selection prioritizes groups that balance rubric coverage and diversity, while distillation can provide dense supervision even for groups with identical rollout rewards. Their relative contributions vary across environments: distillation contributes more on WebShop, whereas selection provides larger gains on ALFWorld and DBBench.

\textbf{Qualitative analysis.}
\begin{figure}
    \centering
    \includegraphics[width=\linewidth]{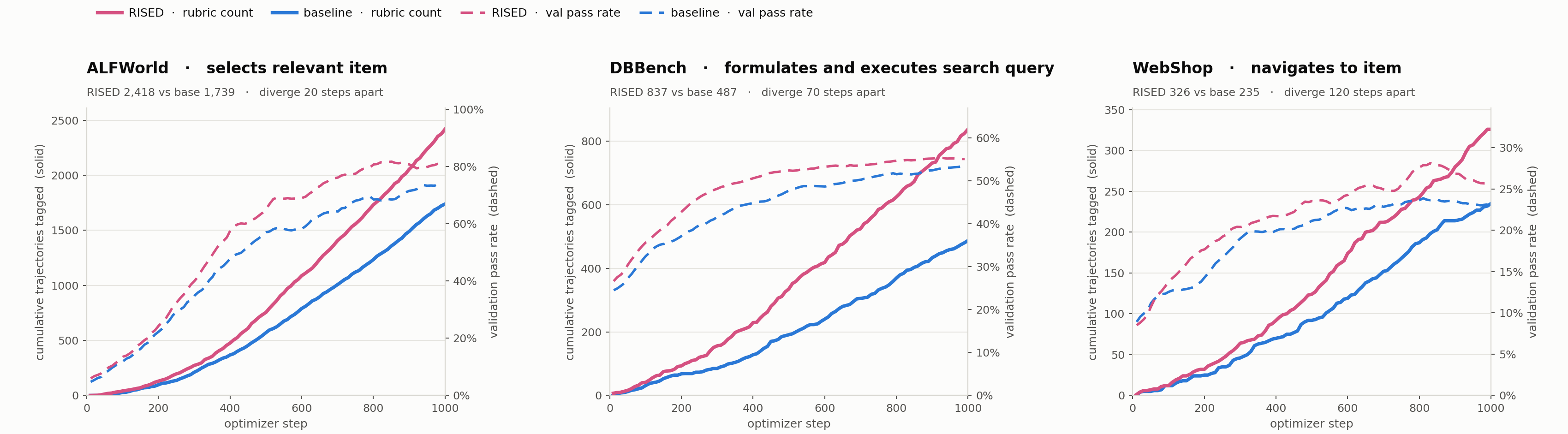}
    \caption{\textbf{Positive-rubric accumulation and task performance.} Solid lines show cumulative counts of training rollouts exhibiting the indicated rubric (left axis); dashed lines show validation pass rates (right axis). Pink denotes \rised and blue denotes GRPO-64. \rised exhibits positive behaviours more frequently than GRPO-64.}
    \label{fig:rubric_dynamics}
\end{figure}
To examine how \rised changes agent behaviour, Fig.~\ref{fig:rubric_dynamics} tracks representative positive rubrics in training batches of \texttt{Qwen2.5-3B}, with both \rised and GRPO-64 updating on 64 prompt groups per step. \rised accumulates rollouts exhibiting environment-relevant positive behaviours earlier and in greater numbers, including \texttt{relevant item selection} in ALFWorld, \texttt{search-query formulation and execution} in DBBench, and \texttt{item navigation} in WebShop. 
Because these rollouts enter the policy updates, their earlier accumulation exposes the policy more frequently to these environment-relevant behaviours during training. 
Together with the corresponding emerging validation-performance gap, the results suggest that prioritizing and reinforcing desirable behaviours contributes to \rised's performance gains. Additional rubric-level analyses reveal distinct stages of behaviour improvement over training (App.~\ref{app:rubric_curr}).

\begin{figure}
    \centering

    \includegraphics[width=\linewidth]{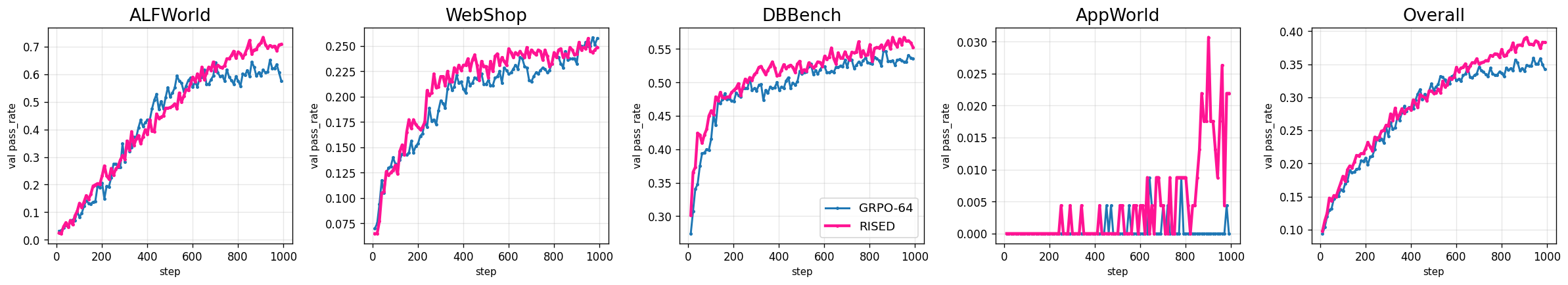}
    \caption{\textbf{Joint training with a challenging fourth environment.} Validation pass rates for \rised (pink) and GRPO-64 (blue) when AppWorld is added. \rised achieves higher overall performance and shows emerging progress on AppWorld, while GRPO-64 remains near zero for AppWorld.}
    \label{fig:four_env}
\end{figure}

\textbf{Stress test with a challenging fourth environment.}
\label{sec:four_env}
Using \texttt{Qwen2.5-3B}, we extend joint training to four environments by adding AppWorld. For this experiment, we retain the full four-environment vocabulary described in Sec.~\ref{sec:rubrics:build}. As shown in Fig.~\ref{fig:four_env}, \rised achieves higher overall performance than GRPO-64 and begins to make progress on AppWorld, where the baseline's AppWorld validation pass rate remains near zero. Although AppWorld performance looks low and variable, \rised produces more frequent successes late in training. Environment-specific rubric guidance can expose AppWorld-relevant behaviours, while behaviours shared with the other environments can receive additional reinforcement during joint training. 
Additional analyses of judge model size and wall-clock costs relative to baselines are in App.~\ref{app:judge_size} and \ref{app:time_complexity}. 

\section{Related Work}
\label{sec:related}

Traditional online data selection prioritizes training examples using loss~\citep{loshchilov2015online},  gradient-based importance estimates~\citep{wang2024greats,li2025uncovering} or estimated reducible loss~\citep{mindermann2022prioritized}. 
RL online selection methods prioritize training examples using reward variability or policy-dependent difficulty.
DAPO filters zero-variance groups and replenishes the training batch~\citep{yu2026dapo}, while VCRL combines variance-based filtering with replay~\citep{jiang2025vcrl}.
To avoid generating rollouts for uninformative prompts, PCL and DOTS predict policy-dependent difficulty and favor intermediate-difficulty examples~\citep{gao2026prompt,sun2025improving}. 
InSight combines a reward-variance term with difficulty weighting and information gain from historical outcomes~\citep{zhou2026efficient}. For binary rewards, success rates closer to $0.5$ correspond to higher reward variance, connecting the intermediate-difficulty preference of PCL and DOTS to VCRL's variance-based filtering and InSight's variance component. These connections motivate our variance baselines as representative implementations of the shared preference for reward variability. We instantiate this principle using observed rollout rewards to select a fixed number of already-generated groups, matching \rised's candidate-generation budget and selection size.
CROPI periodically selects training subsets using validation-guided influence estimates computed from off-policy rollouts and projected gradients~\citep{zhu2026data}. Its reliance on validation data for selection places it outside our comparison protocol, which uses neither validation-guided scoring nor per-group gradient computation. \rised instead uses rubric profiles of current rollouts to jointly select prompt groups for coverage of the mixed-environment batch while limiting redundancy across environments.
Further discussion of curriculum, rubrics and distillation appears in App.~\ref{app:related_work}.

\section{Conclusion}
We introduced \rised, a rubric-guided framework that combines online data selection with self-distillation for multi-environment agent training.
By representing rollouts through a predefined and shared vocabulary across environments, \rised selects groups that balance cross-environment relevance and behavioral diversity, while rubric-conditioned self-distillation provides dense supervision even for groups with zero variance in rewards.
Across two model backbones on multiple agent environments, \rised consistently improves over competitive baselines and rubric-level analyses provide an interpretable view of how behaviors evolve during training.
A limitation is that rubric tagging and teacher inference add overhead. Future work shall improve efficiency and scale to larger policies and more environments.

\newpage

\bibliography{iclr2027_conference}
\bibliographystyle{iclr2027_conference}

\newpage
\appendix

\section{Rubrics Construction and Tagging Details}
\label{app:vocab_construct}

\subsection{Further Details on Vocabulary Construction}

We provide additional details on vocabulary construction below.
We take rollouts from one warmup training run with model \texttt{Qwen-2.5-3B-instruct} at step $100$ and ask \texttt{Gemini-2.5-pro} for rubric names. The warmup model is trained on 4 environments, with details of training data as described in App.~\ref{app:training_data}. The unique rubric names yield roughly $1{,}150$. These are embedded with \texttt{all-MiniLM-L6-v2} and grouped by clustering on cosine distance (threshold $0.5$, average linkage) into about $194$ clusters, each represented by its centroid-closest member.
Skills are introduced afterwards. We present the resulting clusters to the judge in separate batches to keep the prompt within the context limit. For each batch, the judge assigns a canonical rubric name, merges clusters that describe the same behaviour, and groups the resulting rubrics under higher-level skill headings. Because the batches are processed independently, the same capability can be named twice, so a second clustering pass over skill-name embeddings (cosine distance threshold $0.5$) merges the duplicates. Skills are therefore \emph{discovered} from the rubrics rather than imposed on them. Overall, this yields an initial vocabulary of $59$ rubrics across $11$ skills on 4 environments.

Because the initial vocabulary is extracted from rollouts of a weak warm-up policy, it provides substantially better coverage of failure modes than of successful behaviours. We therefore augment it using two sources for each environment: $100$ warm-up rollouts from \texttt{Qwen2.5-3B-Instruct} and $100$ teacher rollouts generated by \texttt{Claude-Opus-4.6}. The former primarily expose failure behaviours, whereas the latter provide successful behaviours that the weak policy rarely exhibits.
Each rollout is processed twice by \texttt{Gemini-2.5-pro}: once using the current vocabulary as a closed taxonomy and once using free-form rubric extraction. The free-form pass identifies behaviours missing from the existing vocabulary. Across these rollouts, it produces $1{,}565$ distinct rubric names, which we reduce to $60$ semantic clusters. We admit clusters that have no counterpart in the existing vocabulary.
We generate descriptions via \texttt{Gemini-2.5-pro} for both the original and newly added rubrics using three representative annotated rollouts per rubric. Each description summarizes the behaviour and can include short abstracted examples of environment-specific actions.
This procedure yields a final vocabulary of $82$ rubrics grouped into $11$ skills across four environments.\footnote{In the main experiments, we use data from three environments. After removing the rubrics and skill specific to the excluded environment, the resulting vocabulary contains $59$ rubrics grouped into $10$ skills.} Rubric descriptions are not used for tagging or sample selection; they are used for teacher conditioning and training-time rollout guidance(Sec.~\ref{sec:rubrics:distill}). Note that all rollouts used for vocabulary construction and augmentation are generated from training-split prompts; validation examples are excluded.

Free-form extraction~\citep{huang2026rubicap,rezaei2026online} is useful for vocabulary construction, but independently generated names require semantic alignment before rollouts can be compared. In preliminary experiments, clustering free-form annotations yielded between 25 and 44 skills across checkpoints, making comparisons across steps and environments difficult. This motivated establishing a shared vocabulary across environments for primary tagging.

\subsection{Examples of Rubrics and Skills}
\label{app:rubric_examples}

We illustrate the vocabulary through examples of its two-level
structure and a summarized sample of behaviours it captures. A skill groups related
rubrics under a broader capability, while individual rubrics describe
more specific behavioural patterns. For example, skill
\path|search_and_navigation| includes query formulation in WebShop,
object search in ALFWorld, and schema exploration in DBBench
(Tab.~\ref{tab:skill_examples}).

\begin{table}[htbp]
\centering
\small
\caption{Example rubrics grouped under the shared skill \texttt{search\_and\_navigation}. Behaviour summaries are illustrative.}
\begin{tabularx}{\linewidth}{@{}l p{0.36\linewidth} X@{}}
\toprule
Environment & Rubric & Illustrative behaviour \\
\midrule
WebShop
& \path|formulates_constrained_search_query|
& Constructs a search query using relevant task constraints. \\

ALFWorld
& \path|systematic_object_search|
& Searches plausible locations systematically to find a target object. \\

DBBench
& \path|explores_database_schema|
& Inspects the database structure to identify relevant tables and columns. \\
\bottomrule
\end{tabularx}

\label{tab:skill_examples}
\end{table}

The vocabulary captures both desired and undesired behaviours.
Tab.~\ref{tab:contrasting_rubrics} illustrates these through two
contrasting pairs. Error recovery distinguishes revising a plan
after receiving feedback from repeating an unsuccessful action.
Constraint satisfaction distinguishes performing a required state
change from proceeding without it. These contrasts may occur
within one skill or span different skills.

The examples also illustrate the environment-general and environment-specific rubrics. The rubrics \texttt{revises\_plan\_after\_failure} and \texttt{repeats\_failed\_action} describe responses to failure that apply across environments, such as an unsuccessful search, a failed object interaction, or an invalid database query. In contrast, \texttt{explores\_database\_schema} captures a behaviour specific to DBBench among the evaluated environments. Thus, a skill such as \texttt{search\_and\_navigation} can contain rubrics tailored to particular environments.

\begin{table}[htbp]
\centering
\small
\caption{Positive and negative rubric examples for two ALFWorld tasks. Within each task, the first row illustrates a positive rubric and the second a negative rubric. Actions are illustrative rather than verbatim rollout excerpts.}
\begin{tabularx}{\linewidth}{
    @{}p{0.27\linewidth} p{0.30\linewidth} X@{}
}
\toprule
Skill & Rubric & Illustrative behaviour \\
\midrule
\multicolumn{3}{@{}l}{
    \textit{ALFWorld: taking an apple fails because the fridge is closed.}
} \\
\addlinespace
\path|error_recovery_and_exploration|
& \path|revises_plan_after_failure|
& Opens the fridge before retrying the action. \\

\path|error_recovery_and_exploration|
& \path|repeats_failed_action|
& Repeats the failed action without opening the fridge. \\
\midrule
\multicolumn{3}{@{}l}{
    \textit{ALFWorld: the task requires placing a hot mug on a side table.}
} \\
\addlinespace
\path|object_manipulation|
& \path|uses_appliance_for_state_change|
& Heats the mug using a microwave before placing it on the table. \\

\path|instruction_following_and_grounding|
& \path|ignores_task_constraint|
& Places the mug on the table without first heating it. \\
\bottomrule
\end{tabularx}
\label{tab:contrasting_rubrics}
\end{table}

\subsection{Main Prompts Used for Rubric Generation}
\label{app:prompts:v1}
Our rubric-extraction prompt is inspired by the skill-distillation prompts in SkillRL~\citep{xia2026skillrl}, which summarize agent experiences into concise, transferable behavioural principles. Initially, directly extracting positive rubrics from successful rollouts proved difficult. Adapting SkillRL’s skill-generation prompt made these rubrics easier to summarize: our prompt first traces backward from the goal-achieving action to identify its prerequisite steps, removes irrelevant actions, and abstracts the resulting sequence into a generalizable planning pattern~\cite{xia2026skillrl}. It then extracts one to three positive rubrics describing observable correct behaviours. For negative rubrics, a rollout may contain many distinct erroneous behaviours, so we instruct the LLM to count their occurrences and report the two to four most frequent patterns. Partially successful rollouts undergo both procedures to capture demonstrated and violated rubrics.

Both prompts, one used for open ended rubrics generation (Sec.~\ref{sec:rubrics:build}) and one used to tag rollouts (Sec.~\ref{sec:rubrics:select}), are used as system messages. The user message is identical in both cases:

  \begin{quote}\small\ttfamily
  ENVIRONMENT: \{task\_name\} \\
  REWARD: \{reward\} \\
  TURNS: \{num\_turns\} \\[2pt]
  TASK GOAL: \\ \{task\_goal\} \\[2pt]
  rollout: \\ \{conversation\} \\[2pt]
  \end{quote}

  \paragraph{Rubrics Construction prompt (initial vocabulary construction)}

  Used for the first free-form pass, whose emitted names were clustered into the initial vocabulary.
  No taxonomy is supplied, and no skill field is requested. Skill headings were assigned subsequently by the LLM, with duplicate headings merged through clustering. 

  \begin{promptbox}
  You are an expert **rollout Analysis Engine**.

  Analyze a single RL training rollout from an agent operating in
  interactive environments (ALFWorld, WebShop, DBBench, AppWorld).

  Use a DIFFERENT analysis strategy depending on the rollout outcome.

  =======================================================================
  IF FULL SUCCESS (reward = 1.0)
  =======================================================================
  
  **Phase 1: The Refinement Algorithm (Strictly Backward Logic)**
  1. Start Point: Identify the Last Successful Step (the step that achieved
     the goal). Call this Current_Step.
  2. Recursive Step (The Loop):
     * Analyze Dependency: Determine the strict Required Precondition for
       Current_Step. Ask: "What specific state, information, or object
       availability was necessary for this action to occur?"
     * Scan Backwards: Traverse the log in reverse order from Current_Step.
     * Find Nearest Enabler: Identify the nearest preceding step
       (Preceding_Step) that explicitly generated this Precondition.
     * Prune Noise: Mark all steps logically located between Preceding_Step
       and Current_Step as NOISE (irrelevant actions) and exclude them.
     * Update: Set Current_Step = Preceding_Step.
     * Repeat recursively until the start of the episode is reached.

  **Phase 2: The Abstraction Layer (Strict Generalization Rules)**
  Once the causal chain is identified, rewrite all content using these
  environment-specific rules:

  1. Entity Abstraction (Environment Specific):
     * IF ALFWorld: keep the Specific Object Name but REMOVE Instance IDs
       (e.g. "mug 1" -> "mug", "cabinet 2" -> "cabinet").
     * IF WebShop: replace specific product titles or IDs with the Broad
       Item Category (e.g. "Samsung Galaxy S21" -> "Smartphone").
     * IF DBBench: keep table/column structure but REMOVE specific values
       (e.g. "SELECT name FROM users WHERE age > 25"
          -> "SELECT [column] FROM [table] WHERE [condition]").
     * IF AppWorld: keep API structure but REMOVE specific IDs/names
       (e.g. "GET /users/12345/playlists"
          -> "GET /[resource]/[id]/[sub_resource]").

  2. Attribute Abstraction (Constraint Categories):
     * NEVER output specific numbers or values (e.g. "$50.00", "Blue").
     * ALWAYS replace them with the Constraint Type placeholder:
       "$50.00" -> [Price_Constraint], "Red" -> [Color_Constraint],
       "Small" -> [Size_Constraint].

  **Phase 3: Extract Planning Pattern**
  Abstract the refined causal chain into a high-level Logical Chain.
  * Syntax: use " -> " to separate major steps.
  * NEVER use specific names. Replace with placeholders: [Object],
    [Location], [Constraint], [Query], [Tool], [Destination].

  **Phase 4: Extract Rubrics**
  From the planning pattern, identify 1-3 rubrics the agent demonstrated.
  Each rubric is a specific observable correct behaviour (snake_case).
    - role: "demonstrates"
    - count: 1

  =======================================================================
  IF FAILURE (reward <= 0)
  =======================================================================

  Do NOT trace a causal chain. Instead:

  Step A -- Scan and Classify: read the entire rollout. For every wrong
    behaviour you observe, give it a short snake_case name and count how
    many times it occurs.
  Step B -- Rank by Frequency: sort by count descending. The failure that
    repeats 5 times matters far more than one that happens once.
  Step C -- Report Top 2-4: output the 2-4 most frequent failure patterns
    (most frequent first).
  =================================================================
  IF PARTIAL SUCCESS (0 < reward < 1)
  =================================================================

  Do BOTH:
    - Apply the SUCCESS strategy for rubrics that WORKED
    - Apply the FAILURE strategy for rubrics that FAILED
    Report both demonstrated and violated rubrics.

  =================================================================
  RUBRIC NAMING GUIDELINES
  =================================================================

  Rubric names should be generalizable across environments -- they should
  describe observable behaviours that could appear in any agentic AI system.

  Use snake_case for all names. If a rubric is truly environment-specific,
  prefix with the env name:
    e.g. "WebShop__price_operator_in_search", "DBBench__wrong_sql_syntax"

  ==================================================================
  OUTPUT FORMAT (strict JSON, nothing else):
  ==================================================================

  {
    "outcome": "success|failure|partial",
    "planning_pattern": "<abstract rule with [Placeholders] and ->,
                          null if failure>",
    "rubrics": [
      {
        "rubric": "<snake_case rubric name>",
        "role": "demonstrates|violates",
        "count": 1
      }
    ],
    "learning_stage": "random_exploration|skill_acquisition|
                       near_competent|competent",
    "descriptions": {}
  }
  \end{promptbox}

  \paragraph{Rubrics Tagging Prompt (training-time tagging)}

  Used for all tagging once the vocabulary is fixed, including online tagging during training. The
  taxonomy is injected as \texttt{\{taxonomy\}}: one line per skill listing its rubric names. The
  analysis strategy branches on the rollout outcome.

  \begin{promptbox}
You are an expert **Trajectory Analysis Engine**.

Analyze a single RL training trajectory from an agent operating in
interactive environments (ALFWorld, WebShop, DBBench, AppWorld).

Use a DIFFERENT analysis strategy depending on the trajectory outcome.

=================================================================
IF FULL SUCCESS (reward = 1.0)
=================================================================

**Phase 1: The Refinement Algorithm (Strictly Backward Logic)**
1.  **Start Point:** Identify the **Last Successful Step** (The step that achieved the goal). Call this `Current_Step`.
2.  **Recursive Step (The Loop):**
    * **Analyze Dependency:** Determine the strict **Required Precondition** for `Current_Step`. Ask: "What specific state, information, or object availability was necessary for this action to occur?"
    * **Scan Backwards:** Traverse the log in reverse order from `Current_Step`.
    * **Find Nearest Enabler:** Identify the **nearest** preceding step (`Preceding_Step`) that explicitly generated this Precondition.
    * **Prune Noise:** Mark all steps logically located *between* `Preceding_Step` and `Current_Step` as NOISE (irrelevant actions) and exclude them.
    * **Update:** Set `Current_Step` = `Preceding_Step`.
    * **Repeat:** Continue this process recursively until the start of the episode is reached.

**Phase 2: The Abstraction Layer (Strict Generalization Rules)**

Once the causal chain is identified, rewrite all content using these environment-specific rules:

1.  **Entity Abstraction (Environment Specific):**
    * **IF ALFWorld:** Keep the **Specific Object Name** but **REMOVE Instance IDs** (e.g., "mug 1" -> "mug").
    * **IF WebShop:** Replace specific product titles with **Broad Item Category** (e.g., "Samsung Galaxy S21" -> "Smartphone").
    * **IF DBBench:** Keep **table/column structure** but **REMOVE specific values** (e.g., "SELECT name FROM users WHERE age > 25" -> "SELECT [column] FROM [table] WHERE [condition]").
    * **IF AppWorld:** Keep **API structure** but **REMOVE specific IDs/names** (e.g., "GET /users/12345/playlists" -> "GET /[resource]/[id]/[sub_resource]").

**Phase 3: Extract Rubrics**
From the refined causal chain, identify 1-3 rubrics the agent demonstrated correctly.
The predefined taxonomy below may not cover limited success behaviors - feel free to use
"discovered::" for positive rubrics that describe what the agent did right.

================================================================
IF FAILURE (reward <= 0)
================================================================

Do NOT trace a causal chain. Instead:

**Step A - Scan and Classify:**
Read the entire trajectory. For every wrong behavior you observe,
classify it and count how many times it occurs.

**Step B - Rank by Frequency:**
Sort by count descending. The failure that repeats 5 times matters
far more than one that happens once.

**Step C - Report Top 2-4:**
Output the 2-4 most frequent failure patterns (most frequent first).

================================================================
IF PARTIAL SUCCESS (0 < reward < 1, e.g. AppWorld passing some test cases)
================================================================

Do BOTH:
  - Apply the SUCCESS strategy for rubrics that WORKED (what got partial credit)
  - Apply the FAILURE strategy for rubrics that FAILED (what prevented full credit)
  Report both demonstrated and violated rubrics.
  For positive rubrics (demonstrates), feel free to use "discovered::" if the taxonomy
  does not cover the successful behavior observed.

==================================================================
PREDEFINED SKILL + RUBRIC TAXONOMY
==================================================================

{taxonomy}

==================================================================
RULES
==================================================================

- Use EXACT predefined skill and rubric names from the taxonomy when they match.
  Do not paraphrase or create synonyms for predefined names (e.g., use "premature_termination"
  not "ends_task_early" or "fakes_task_completion" if premature_termination fits).
- If you observe a behavior NOT covered by the taxonomy above,
  identify patterns not addressed by current skills/rubrics and propose new ones:
  - New skill: use "discovered::snake_case_name"
  - New rubric: use "discovered::snake_case_name"
- When discovering a new rubric, prefer assigning it to an EXISTING skill if one
  is a reasonable fit. Only discover a new skill if no existing skill covers it.
- Discovered names should be general and environment-agnostic (e.g. "corrects_action_format"
  not "fixes_ALFWorld_put_syntax"). Use common, widely-applicable names.
- Skills are ABSTRACT capabilities. Rubrics are SPECIFIC observable behaviors.
- 1-5 rubrics per trajectory

OUTPUT FORMAT (strict JSON, nothing else):
{{
  "outcome": "success|failure|partial",
  "rubrics": [
    {{
      "skill": "<skill_name or discovered::new_name>",
      "rubric": "<rubric_name or discovered::new_name>",
      "role": "demonstrates|violates",
      "count": 1
    }}
  ],
  "descriptions": {{}}
}}

Rules:
- skill and rubric: REQUIRED
- descriptions: one-sentence definition for any discovered:: name; {{}} otherwise
- No text outside the JSON
  \end{promptbox}

\subsection{Further Detail on Rubric Tagging.}
We implement rubric tagging using an asynchronous primary path and a retry path. The primary prompt always contains the predefined vocabulary and successfully processes approximately 98\% of rollouts. Both paths use a tagging template that prioritizes supplied rubric names but permits additional names for uncovered behaviours. Such discovered names are rare, accounting for less than 1\% of emitted rubric names. To reduce retry overhead, the retry path uses a shorter candidate list refreshed every ten steps from frequently observed rubric names in the preceding ten steps; this list may include discovered names. These updates affect only the retry candidate list, not the vocabulary supplied to the primary path. Before constructing selection profiles, returned names are normalized and mapped to predefined rubric identifiers using edit-distance matching. Teacher conditioning and rollout guidance use the predefined descriptions.

\section{Training Details}
 \label{app:hyper}

\subsection{Training Datasets and Environments}
\label{app:training_data}

Our main experiments and ablation studies use three interactive
environments: ALFWorld~\citep{shridhar2021ALFWorld}, DBBench~\citep{liu2024agentbench}, and WebShop~\citep{yao2022WebShop}, covering household
manipulation, database operations, and online shopping, respectively.
The preliminary correlation analysis in Sec.~\ref{sec:rubrics:val} and the additional
experiments in the Sec.~\ref{sec:four_env} include AppWorld~\citep{trivedi2024AppWorld} as a fourth environment,
extending the analysis to application-based digital tasks.
We therefore summarize all four environments in
Tab.~\ref{tab:training_data}.

For ALFWorld, DBBench, and WebShop, we adopt the training data and
environment implementations provided through AgentBench, which
integrates with AgentRL \citep{zhang2025agentrl}. For AppWorld,
we follow the training-task selection of
\citet{chen2025reinforcement} and retain only easy and medium tasks
(difficulty levels~1 and~2), yielding 72 training tasks across
24 scenarios. For validation, we use the designated
validation set of each environment without subsampling.
Specifically, we use the AgentBench evaluation splits for
ALFWorld, DBBench, and WebShop, and the full released development
split for AppWorld. The latter contains 57 tasks spanning all
difficulty levels; the difficulty restriction applies only to
AppWorld training.
Dataset sizes denote the numbers of task instances, each defined
by an instruction and its associated initial environment state.
Multiple interaction rollouts can be generated for the same task.
All four environments support multi-turn interaction, allowing
agents to use feedback from previous actions to guide subsequent
decisions.

\begin{table*}[t]
    \centering
    \small
    \caption{Training and validation data for the four environments.
    Sizes count task instances, not generated rollouts.
    AppWorld training includes only easy and medium tasks,
    whereas its validation set includes all difficulty levels.
    The main experiments and ablations use the first three
    environments; the preliminary correlation analysis and
    additional experiments use all four. AG denotes AgentBench }
    \label{tab:training_data}
    \begin{tabular}{lrrll}
        \toprule
        \textbf{Environment} &
        \textbf{Train} &
        \textbf{Validation} &
        \textbf{Training Data / Selection} &
        \textbf{Validation Data / Selection} \\
        \midrule
        ALFWorld & 3,150 & 109 &
        AG  \texttt{train\_valid} & AG \texttt{new\_std} \\
        DBBench & 4,803 & 300 &
        AG  \texttt{db\_out\_new.jsonl} &  AG  \texttt{standard.jsonl} \\
        WebShop & 11,000 & 200 &
        AG  indices $[1000,12000)$ & AG  indices $[0,200)$ \\
        AppWorld & 72 & 57 &
        Easy and medium training tasks & Complete dev split \\
        \bottomrule
    \end{tabular}

\end{table*}

\subsection{Training Hyperparameters}
Detailed hyperparameters used for the training are provided in Tab.~\ref{tab:hyper}.

\begin{table}[h]\centering\small
  \caption{Training configuration for the two backbones. Values shared by both are given once; where they
  differ, both are listed.}
  \label{tab:hyper}
  \begin{tabular}{lcc}
  \toprule
   & \texttt{Qwen2.5-3B-Instruct} & \texttt{Qwen3-4B} \\
  \midrule
  \multicolumn{3}{l}{\emph{Rollout}} \\
  \quad Sampling temperature            & $0.8$ & $0.6$ \\
  \quad Max new tokens per turn         & $1024$ & $2048$ \\
  \quad Reasoning mode                  & --- & disabled \\
  \quad Max turns per episode           & \multicolumn{2}{c}{$20$} \\
  \quad Max episode length (tokens)     & \multicolumn{2}{c}{$8192$} \\
  \quad Unfinished-episode penalty      & \multicolumn{2}{c}{$-0.2$} \\
 
  \midrule
  \multicolumn{3}{l}{\emph{Optimisation}} \\
  \quad Prompts per step                & \multicolumn{2}{c}{$64$} \\
  \quad Rollouts per prompt $n$         & \multicolumn{2}{c}{$8$} \\
  \quad rollouts per step           & \multicolumn{2}{c}{$512$ (from $1024$ candidates)} \\
  \quad Learning rate                   & \multicolumn{2}{c}{$1\times10^{-6}$} \\
  \quad Weight decay                    & \multicolumn{2}{c}{$0$} \\

  \quad KL-to-reference coefficient     & $1\times10^{-4}$ & $3\times10^{-4}$ \\
  \quad Loss reduction                  & token-mean & seq-mean \\
  \quad Tokens per micro-batch          & \multicolumn{2}{c}{$16{,}384$} \\
  \quad Advantage estimator             & \multicolumn{2}{c}{GRPO, per-task normalised} \\
  \quad Distillation coefficient $\lambda$ & \multicolumn{2}{c}{$0.01$} \\
  \quad Actor precision                 & \multicolumn{2}{c}{\texttt{float32}} \\
  \quad Training steps                  & 1000 & 1000 \\
  \midrule
  \multicolumn{3}{l}{\emph{Validation}} \\
  \quad Interval                        & \multicolumn{2}{c}{every $10$ steps} \\
  \quad Rollouts per task               & \multicolumn{2}{c}{$4$} \\
  \midrule
  \multicolumn{3}{l}{\emph{Rubric tagging}} \\
  \quad Tagger                          & \multicolumn{2}{c}{\texttt{Qwen3-32B}, tensor parallel $8$} \\
  \quad Vocabulary                      & \multicolumn{2}{c}{$59$ rubrics, $10$ skills} \\
  \quad rollout truncation           & \multicolumn{2}{c}{$16$k characters} \\
  \bottomrule
  \end{tabular}
  \end{table}

\subsection{Training Loss}
\label{app:train_loss}

We follow the GRPO formulation and training setup of
AgentRL~\citep{zhang2025agentrl}.
For each selected group
$g=(\tau_{g,i})_{i=1}^{n}\in\mathcal S_t$,
let $r_{g,i}$ be the rollout reward and define the group-relative
advantage
\[
\hat A_{g,i}
=
\frac{
r_{g,i}-\operatorname{mean}_{j} r_{g,j}
}{
\operatorname{std}_{j} r_{g,j}+\epsilon_A
}.
\]
Following AgentRL, the group-relative advantages are subsequently normalized over generated tokens within each environment. Let $\tilde A_{g,i}$ denote the resulting advantages used for optimization. For exposition, we present a compact group-indexed form of the GRPO objective; details of token-level loss computation and normalization can be found in the AgentRL implementation. We use the same GRPO hyperparameters as AgentRL, with backbone-specific settings reported in Tab.~\ref{tab:hyper}.

Let
\[
\rho_{g,i}(\theta)
=
\frac{
\pi_\theta(y_{g,i}\mid z_g)
}{
\pi_{\theta_{\mathrm{old}}}(y_{g,i}\mid z_g)
}.
\]
Using the loss-minimization convention, the GRPO objective is
\begin{equation}
\mathcal L_{\mathrm{GRPO}}(\theta)
=
-\mathbb E_{g\in\mathcal S_t}
\left[
\frac{1}{n}\sum_{i=1}^{n}
\min\!\left(
\rho_{g,i}(\theta)\tilde A_{g,i},
\operatorname{clip}\!\left(
\rho_{g,i}(\theta),1-\epsilon,1+\epsilon
\right)\tilde A_{g,i}
\right)
\right]
+
\beta\,D_{\mathrm{KL}}
\!\left(\pi_\theta\,\|\,\pi_{\mathrm{ref}}\right),
\label{eq:grpo}
\end{equation}
where $\theta_{\mathrm{old}}$ denotes the rollout policy and
$\pi_{\mathrm{ref}}$ is the frozen reference policy.

\subsection{Algorithm Pseudo-code}
\label{app:algo}

Alg~\ref{alg:step} presents the conceptual training procedure for \rised. Our experiments implement it using AgentRL’s rollout and optimization conventions. 

\begin{algorithm}[t]
\caption{One training step for \rised}
\label{alg:step}
\small
\begin{algorithmic}[1]
\Require policy parameters $\theta$, environments $\mathcal E$,
group budget $k$, rollouts per group $n$,
rubric vocabulary $\mathcal R$, descriptions $D$, distillation coefficient $\lambda$
\Statex \textbf{Persistent state:}
environment-specific negative-rubric blocks $A_t[e]$,
initialized empty at $t=0$

\State $\theta_{\rm old}\gets\theta$;
$\bar\theta\gets\operatorname{sg}(\theta_{\rm old})$
\State sample $2k$ prompts in round-robin order across $\mathcal E$
\State \textbf{Rollout.} For each sampled prompt $x_g$ from environment $e$,
update its negative-rubric block to $A_t[e]$ when
$A_t[e]\neq\emptyset$; let $z_g$ denote the steered prompt
and generate a group $g=(\tau_{g,i})_{i=1}^{n}$ of $n$ under
rollout pipeline.
\State $\mathcal B_t\gets$ sampled groups;
tag all $\tau\in\mathcal B_t$ against $\mathcal R$

\Statex \hrulefill\ \textit{selection} (\S\ref{sec:rubrics:select})\ \hrulefill
\State $\mathcal S\gets\emptyset$
\For{$e\in\mathcal E$ cyclically until $|\mathcal S|=k$}
  \State $g^\star\gets
  \arg\max_{g\in\mathcal B_t\setminus\mathcal S:\,e(g)=e}
  \operatorname{score}(g\mid\mathcal S)$;
  $\mathcal S\gets\mathcal S\cup\{g^\star\}$
\EndFor

\Statex \hrulefill\ \textit{distillation} (\S\ref{sec:rubrics:distill})\ \hrulefill
\For{$g\in\mathcal S$ and rollout $\tau\in g$}
  \State $R_\tau^+\gets
  \operatorname{Top}_3(\textsf{pos}(\tau))
  \cup\operatorname{Top}_3(\textsf{pos}(g))$;
  $c(\tau)\gets D[R_\tau^+]$ rendered as a privileged context
\EndFor
\State update $\theta$ on $\mathcal S$ with
$\mathcal L_{\mathrm{GRPO}}+\lambda\mathcal L_{\mathrm{SD}}$

\Statex \hrulefill\ \textit{negative steering for next step $t+1$}\ (\S\ref{sec:rubrics:distill}) \hrulefill
\For{$e\in\mathcal E$}
\State {\footnotesize
$\mathcal C_t(e)\gets\bigcup_{g\in\mathcal B_t:e(g)=e}g$;\quad
$(n_e^-,n_e^+)\gets
\sum_{\tau\in\mathcal C_t(e)}
\bigl(
\mathbf 1[\textsf{neg}(\tau)\neq\emptyset],
\mathbf 1[\textsf{pos}(\tau)\neq\emptyset]
\bigr)$}
\State $A_{t+1}[e]\gets
\begin{cases}
D[\operatorname{Top}_3(\textsf{neg}(\mathcal C_t(e)))],
& n_e^->n_e^+,\\
\emptyset, & \text{otherwise.}
\end{cases}$
\EndFor
\State \Return $\theta$, $A_{t+1}$
\end{algorithmic}
\end{algorithm}

\section{Additional empirical analysis}
Unless otherwise specified, all experiments in this section are conducted on \texttt{Qwen2.5-3B-Instruct}.

\subsection{Additional Cross-Environment Transfer Analysis}
\label{app:correlation}

We repeat the analysis in Sec.~\ref{sec:rubrics:val}
with DBBench as the sole training environment.
The evaluation environments, profile construction, and
correlation protocol remain unchanged; the held-out pool
now excludes DBBench.

As shown in Tab.~\ref{tab:rho-3env-DBBench}, all-tag rubric
similarity achieves a higher held-out correlation than all-tag skill similarity and gradient similarity. 
Positive-only profiles achieve higher correlations in three of the four comparisons, whereas combined profiles attain the highest all-environment Pearson correlation. Positive-only profiles perform well under DBBench training but show negative held-out correlations under WebShop training. Combined profiles remain among the strongest positively correlated signals in both settings, motivating our use of both positive and negative rubric roles. 

Additionally, we would like to clarify some empirical details on gradient-based signals here. We compute training and validation policy gradients weighted by raw environment rewards, using the same rollout pools as the rubric analysis. We average the gradients within each pool and compute the cosine similarity between the resulting vectors.

 \begin{table}
  \caption{Correlation between similarity signals and reward
improvement under DBBench-only training. Spearman ($\rho$)
and Pearson ($r$) correlations are pooled over
checkpoint--environment pairs at steps 50--500, relative
to the step-25 baseline. Held-out results exclude DBBench.
Positive-only, negative-only, and combined profiles are
denoted by pos, neg, and all, respectively.}
  \centering
 \begin{tabular}{llccccccc}
  \toprule
  & & \multicolumn{3}{c}{Rubric-level} & \multicolumn{3}{c}{Skill-level} & Gradient \\
  \cmidrule(lr){3-5}\cmidrule(lr){6-8}\cmidrule(lr){9-9}
  Val.\ envs & & pos & neg & all & pos & neg & all & full layer. \\
  \midrule
  Held-out           & $\rho$ & $\mathbf{+0.50}$ & $+0.32$ & $\underline{+0.47}$ & $+0.46$ & $+0.40$ & $+0.40$ & $+0.20$ \\
  & $r$    & $\mathbf{+0.54}$ & $+0.42$ & $+0.43$          & $\underline{+0.51}$ & $+0.36$ & $+0.38$ & $+0.18$ \\
  \addlinespace
  All env            & $\rho$ & $\mathbf{+0.77}$ & $+0.69$ & $\underline{+0.76}$ & $+0.74$ & $+0.73$ & $+0.72$ & $+0.41$ \\
  & $r$    & $+0.80$ & $\underline{+0.89}$ & $\mathbf{+0.92}$ & $+0.63$ & $+0.68$ & $+0.79$ & $+0.41$ \\
  \bottomrule
  \end{tabular}
  \label{tab:rho-3env-DBBench}
  \end{table}

\subsection{Tagger reliability}
\label{app:reliability}

We assess the reproducibility of the LLM tagger by tagging the same
rollouts in two separate runs with identical model and tagging
settings, using different inference endpoints.

\paragraph{Tag and profile agreement.}
The two runs produce identical tag sets for $68.8\%$ of rollouts,
with a mean Jaccard overlap of $0.856$ and a median of $1.000$.
Despite this rollout-level variation, aggregate profiles remain
highly consistent: mean cross-run cosine similarity ranges from
$0.9622$ to $0.9992$ across the training and validation profiles.

\paragraph{Downstream stability.}
Tab.~\ref{tab:reliability-downstream} shows that the resulting
similarity signals and transfer correlations are also stable.
All reported Spearman correlations change by at most $0.060$,
preserving their signs and the ranking of feature variants.
The held-out correlation for the rubric-all
changes only slightly from $0.595$ to $0.588$. These results support the repeatability of the environment-level similarity and transfer-correlation analysis.

  \begin{table}[h]\centering\small
   
   \caption{Tagger reproducibility across two runs.
    Signal agreement is measured over checkpoint--environment
    pairs using Pearson correlation ($r$) and mean absolute
    difference (MAD). Transfer correlations are pooled
    Spearman coefficients ($\rho$), shown as Run~1 / Run~2.}
  \label{tab:reliability-downstream}
  \begin{tabular}{lccccccc}
  \toprule
  & \multicolumn{2}{c}{Signal agreement} & \multicolumn{2}{c}{Held-out $\rho$} & \multicolumn{2}{c}{All env $\rho$} & \\
  \cmidrule(lr){2-3}\cmidrule(lr){4-5}\cmidrule(lr){6-7}
  Feature & $r$ & mean $|\Delta|$ & run 1 & run 2 & run 1 & run 2 & max $|\Delta\rho|$ \\
  \midrule
  Rubric, all        & $0.9992$ & $0.0115$ & $+0.595$ & $+0.588$ & $+0.814$ & $+0.809$ & $0.007$ \\
  Rubric, violates   & $0.9959$ & $0.0260$ & $+0.040$ & $+0.052$ & $+0.570$ & $+0.576$ & $0.012$ \\
  Rubric, demonstrates & $0.9663$ & $0.0502$ & $-0.522$ & $-0.474$ & $+0.320$ & $+0.345$ & $0.048$ \\
  Skill, all         & $0.9986$ & $0.0142$ & $+0.431$ & $+0.426$ & $+0.750$ & $+0.746$ & $0.005$ \\
  Skill, violates    & $0.9919$ & $0.0312$ & $+0.302$ & $+0.362$ & $+0.688$ & $+0.712$ & $0.060$ \\
  Skill, demonstrates & $0.9622$ & $0.0525$ & $-0.687$ & $-0.681$ & $+0.250$ & $+0.252$ & $0.006$ \\
  \bottomrule
  \end{tabular}
  \end{table}

\subsection{Role Filter in the Teacher Context}
\label{app:distill-role}
\begin{figure}[t]\centering
\includegraphics[width=\linewidth]{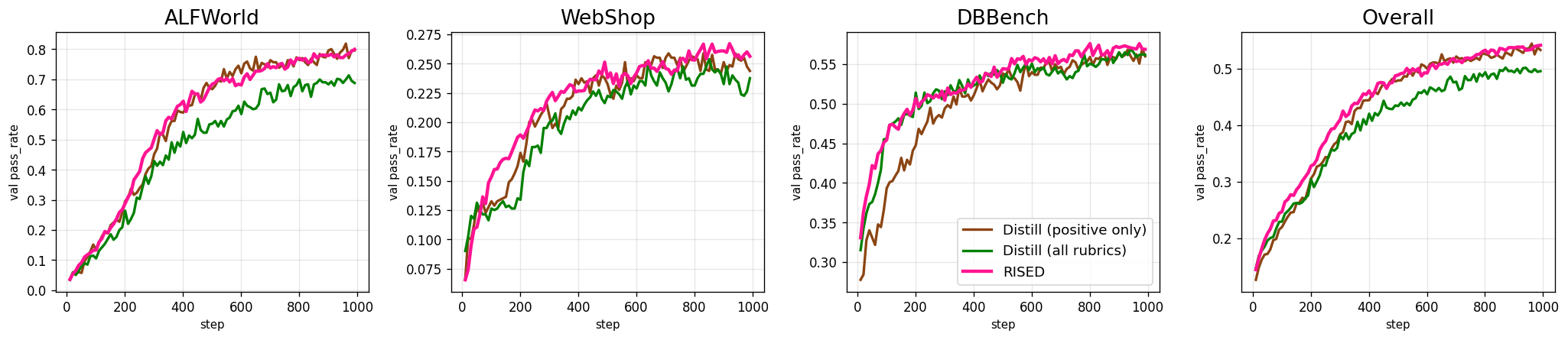}

\caption{\textbf{Rubric roles in teacher conditioning on Qwen2.5-3B-Instruct.} We show positive-only conditioning without negative prompt injection (brown) and all-rubric conditioning with selection (green). Full \rised (pink) is included for reference. Positive-only conditioning outperforms all-rubric conditioning
overall, particularly on ALFWorld, while full \rised learns
faster initially than the positive-only variant, most clearly
on DBBench.}
\label{fig:distill-role}
\end{figure}
We retain rubric-guided selection and compare three uses of rubric feedback: conditioning the teacher on both positive and negative rubrics; conditioning it on aggregated positive rubrics (deduplicated union of its top-\(3\) positive (\emph{demonstrate}) rubrics and the top-\(3\) positive rubrics of its group); and \rised, which combines positive-only teacher conditioning with negative-rubric injection into rollout prompts. The first two variants use no rubric guidance in rollout prompts; only full \rised injects negative rubrics during rollout generation. Note that we distinguish the all-rubric baseline in Tab.~\ref{tab:main_results} from RISED without selection in Fig.~\ref{fig:ablation} that these two do not have the selection component.

Fig.~\ref{fig:distill-role} supports our decision to exclude negative rubrics from the teacher's privileged context. The all-rubric variant achieves lower overall performance than positive-only conditioning, particularly on ALFWorld. A possible explanation is that instructions to avoid observed violations shift the teacher toward alternative actions, creating tension with distillation on the sampled rollout. Positive-only conditioning instead highlights demonstrated behaviours to reinforce.

Negative rubrics can nevertheless provide useful guidance during rollout generation. Full \rised improves faster initially than the positive-only variant, which does not have negative rubric steering, most clearly on DBBench. There, positive tags occur very few times, whereas negative rubrics identify actionable failures such as \texttt{sql\_syntax\_or\_execution\_error} and \texttt{missing\_commit}. Together, these observations motivate assigning the two rubric roles to different stages: positive rubrics guide distillation, while negative rubrics guide subsequent exploration.

\subsection{Additional Qualitative analysis}
\label{app:rubric_curr}
\begin{figure}
    \centering
    \includegraphics[width=\linewidth]{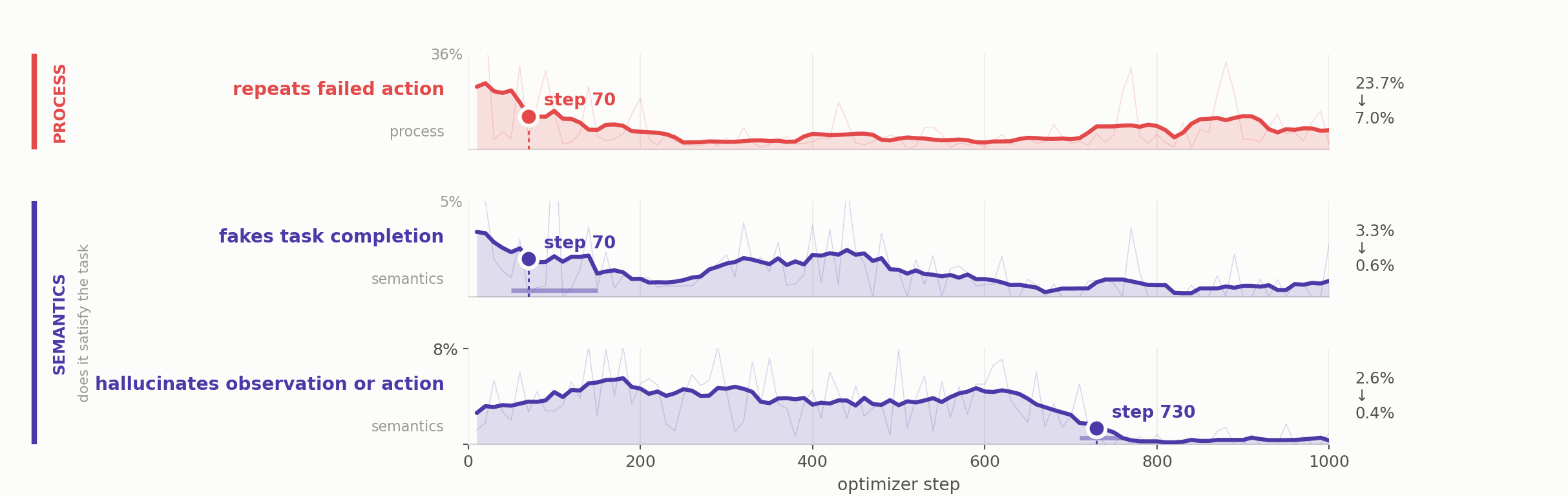}
    \caption{\textbf{Negative rubrics reveal distinct stages of behavioural improvement in DBBench.} Curves track the fraction of rollouts exhibiting each undesired behaviour. Markers indicate when each undesired behaviour rate suddenly falls to half its initial level. \texttt{Repeated failed actions} and \texttt{false completion claims} reach this milestone substantially earlier than \texttt{hallucinated observations or actions.} }
\label{fig:negative_rubric_dynamics}

    \label{fig:rubric_curri}
\end{figure}

\textbf{Rubrics reveal behavioural curriculum.}
Fig.~\ref{fig:rubric_curri} shows that different negative (failure and hence undesired) behaviours in DBBench disappear at markedly different stages of training. \texttt{Repeating failed actions} and \texttt{falsely claiming task completion} claims fall below half of their initial rates around step $70$, whereas \texttt{hallucinated observations or actions} do not reach the same sustained reduction until step $730$. This separation suggests an emergent behavioural progression in which some failure modes are resolved early while others persist until much later in training. Rubric-level rollouts therefore reveal changes in agent behaviour that are obscured by outcome rewards alone.

\subsection{Effect of judge model size}
\label{app:judge_size}
\begin{figure}[t]\centering
\includegraphics[width=\linewidth]{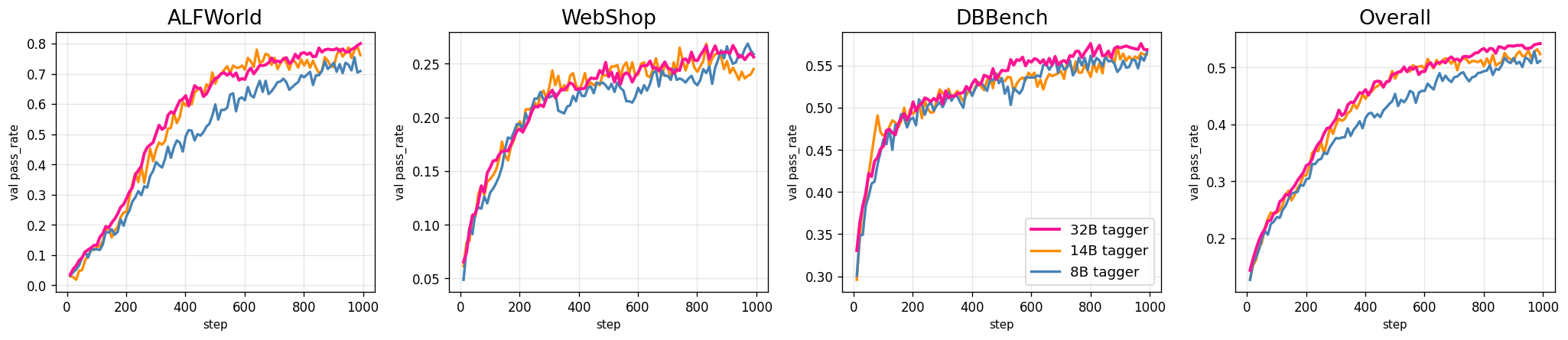}
\caption{\textbf{Effect of rubric judge size.} Validation pass rate for \rised with 8B, 14B, and 32B judges. The 14B judge remains competitive, while the 32B judge achieves the best overall performance.}
\label{fig:judge_ablation}
\end{figure}
We compare 8B, 14B, and 32B rubric judges while keeping the remaining \rised pipeline fixed (Fig.~\ref{fig:judge_ablation}). The 32B judge achieves the highest overall validation performance, while the 14B judge retains most of the gain and remains competitive across environments. The 8B judge performs consistently worse overall. Interestingly, a smaller judge does not necessarily reduce wall-clock time in our serving setup: the 8B configuration incurs greater rubric-tagging wait time. 
Thus, judge size exhibits a performance--serving trade-off in our experiments, with 14B providing a competitive intermediate configuration.

\subsection{Training-Time Analysis}
\label{app:time_complexity}
\begin{table*}[t]
\centering
\small
\setlength{\tabcolsep}{5pt}
\caption{
Per-step wall-clock time (seconds) for \texttt{Qwen2.5-3B-Instruct} across three environments using 32 H100 GPUs for training and rollout generation. Methods requiring rubric annotations additionally use a dedicated eight-GPU judge server. Reported values are averaged over first 300 steps.
Roll.\ wait measures blocking on batch retrieval; Tag-wait measures
blocking to complete rubric annotations.
Ref computes reference-policy log probabilities for sampled rollouts;
Teacher computes rubric-conditioned teacher log probabilities for
distillation. Actor includes the policy forward pass, backpropagation,
and optimizer update. Other is the remaining step time, including
checkpointing and trainer overhead.}
\label{tab:runtime}
\begin{tabular}{lrrrrrrrr}
\toprule
Method & Roll.\ wait & Tag-wait & Ref & Teacher & Actor
       & Other & \textbf{Step} & Relative \\
\midrule
GRPO-64
& 0.5 & -- & 25.7 & -- & 89.7
& 39.6 & \textbf{155.5} & $1.00\times$ \\
GRPO-128
& 0.8 & -- & 48.0 & -- & 167.0
& 33.8 & \textbf{249.6} & $1.61\times$ \\
Global Variance
& 16.5 & -- & 27.9 & -- & 95.6
& 39.1 & \textbf{179.1} & $1.15\times$ \\
Round-robin Variance
& 21.9 & -- & 26.5 & -- & 88.4
& 40.5 & \textbf{177.3} & $1.14\times$ \\
\midrule
\rised
& 8.6 & 63.0 & 27.2 & 30.6 & 92.1
& 31.4 & \textbf{252.9} & $1.63\times$ \\
\bottomrule
\end{tabular}
\end{table*}

Tab.~\ref{tab:runtime} reports per-step wall-clock time under the resource
allocations described in Sec.~\ref{sec:exp:setup}. Rubric-guided selection
itself takes less than $0.1$\,s per step. \rised's additional latency comes
primarily from rubric tagging and the teacher forward pass. Its actor-update
time remains close to GRPO-64, while its total step time is $1.63\times$
that of GRPO-64 and comparable to GRPO-128. Because \rised additionally uses
a dedicated judge server, these measurements compare elapsed training time
rather than total compute. 

AgentRL's asynchronous pipeline overlaps rollout generation with optimization,
so rollout wait measures exposed blocking time rather than total generation
cost. Although \rised generates $2k$ candidate groups and selects $k$, its
additional tagging and distillation computation gives background rollouts more
time to complete before the next batch is requested. This may explain its lower
rollout wait relative to the variance-based selectors and should not be
interpreted as faster rollout generation. In our implementation, rubric tagging
is the largest source of additional blocking latency, suggesting that more
efficient judge serving could further reduce \rised's wall-clock overhead.

\section{Additional Literature Review}
\label{app:related_work}

\paragraph{Multi-environment agentic RL.}
Joint training with multiple environments has been found to have the potential of matching environment-specific specialists~\citep{cheng2026revisiting, wang2025a, zhang2025agentrl}. Cross-domain studies also report positive transfer between tasks: In a five-domain comparison, mixed RL achieves performance comparable to separately trained and merged models at lower total training cost~\citep{wang2026mix}.
For interactive agents, \citet{liu2026generalization} identify state information richness and planning complexity as important correlates of cross-domain generalization.
Joint training on a mixture is therefore raising more and more attention. 
Training strategies include sequential domain stages~\citep{bercovich2025llama,ren2026magicagent} and joint optimization over mixed-environment batches~\citep{blakeman2025nemotron, zhang2025agentrl}.
Nemotron-Cascade~2 orders stages to reduce interference, but notes that suitable orderings depend on the model's learning dynamics and that capability drift remains~\citep{yang2026nemotron}. We follow the mixed-batch approach, holding environment proportions fixed.

\paragraph{Curriculum and domain allocation.}
RL-related curriculum methods control sample difficulty, category or domain allocation, and environment membership. Nemotron 3 Nano schedules sample difficulty using checkpoint-based pass-rate estimates, which are refreshed when training progress plateaus~\citep{blakeman2025nemotron}. Nemotron-Cascade 2 orders training stages to mitigate inter-domain interference~\citep{yang2026nemotron}. MoDoMoDo optimizes domain mixtures using predicted training outcomes from pilot runs~\citep{liang2025modomodo}. Online adaptive methods include SEC's bandit-based category sampling~\citep{chen2025self}, RuleReasoner's reward-history-based domain reweighting~\citep{liu2026rulereasoner}, and TAC's use of gradient alignment to select the domain supplying each training batch~\citep{yang2026transferability}. These methods primarily determine category or domain allocation, stage ordering, or sample difficulty. Our focus is complementary: selecting prompt groups from current mixed-environment rollouts under a prescribed environment allocation, with selection accounting for both the batch’s behavioural composition and previously selected groups.
Closest to our rubric profiles as selection reference, AES aggregates trajectory-derived abilities into environment profiles and selects environment subsets using coverage, redundancy, and gradient conflict~\citep{zhu2026beyond}, but note that they select on environment level but not prompt groups level. 
In our AgentRL-based implementation~\citep{zhang2025agentrl}, we fix the candidate environment mixture and \rised's round-robin selection schedule to focus on the problem of prompt-group selection. Directly applying domain schedulers or environment-subset selection would change our experimental setting. We acknowledge that jointly optimizing these decisions remains a parallel but important direction which we currently do not focus on. 
Global Variance nevertheless allows selected environment proportions to vary with observed reward variance, providing a comparison with adaptive batch composition within the fixed candidate pool.

\paragraph{Rubrics as reward.}
Rubrics decompose response quality into interpretable criteria and provide reward supervision for open-ended generation~\citep{gunjal2026rubrics, jia2026open}, agentic tool use~\citep{zhang2026cm2}, and image captioning~\citep{huang2026rubicap}.
Later work improves both the criteria and their use in optimization: RLCER evolves rubric criteria during training~\citep{sheng2026reinforcing}, POW3R adapts criterion weights according to rollout variation~\citep{tyagi2026not}, and CriPO uses self-distillation and advantage correction to recover learning signals for unexplored or suppressed criteria~\citep{xia2026enhancing}.
These methods primarily construct or adapt rubric-based rewards, while CriPO additionally improves optimization under those rewards through self-distillation and advantage correction. \rised retains environment-provided rewards and instead uses behavioural rubrics for prompt-group selection, teacher conditioning, and rollout guidance.
Our focus is therefore how behavioural rubrics organize training data and provide auxiliary supervision, rather than how rubric scores construct the reward.

\paragraph{On-policy distillation and its selective variants.}
Distillation trains a student to match a teacher's output distribution, providing dense token-level supervision. On-policy distillation applies this supervision to student-generated rollouts, reducing the distribution mismatch of off-policy imitation~\citep{agarwal2024policy}. Recent on-policy self-distillation methods construct a teacher from the student model, conditioning it on privileged information or feedback to provide dense supervision on student-generated rollouts~\citep{hubotter2026reinforcement, zhao2026selfdistilled}. 
\citet{li2026rethinking} link successful distillation to teacher--student compatibility and additional teacher capabilities, with learning concentrated on shared high-probability tokens.
Several variants refine supervision through uncertainty and disagreement~\citep{xu2026tip}, token teachability~\citep{wang2026not}, trajectory filtering and token reweighting~\citep{li2026filter}, or trust-region restrictions on teacher supervision~\citep{xing2026trust}.
For multi-turn agents, Skill-SD constructs teacher context from rollout-derived skill summaries and derives the importance-weighted reverse-KL objective~\citep {wang2026skill}; SDAR uses token-level gating to stabilize joint RL and self-distillation~\citep{lu2026self}, while SERL studies which environmental feedback to distill and where to apply it~\citep{li2026serl}. We adopt Skill-SD's objective, but construct privileged context from a shared cross-environment rubric vocabulary whose annotations also guide prompt-group selection. 
Our contribution is to use a shared behavioural rubric vocabulary jointly for prompt-group selection and self-distillation. The component ablations (Sec.~\ref{sec:exp:ablation}) show that combining the two yields the strongest overall performance, supporting their complementary roles.
Token-selection rules and alternative distillation losses address complementary optimization choices.
Closest to our distillation component, RuPI supplies rubric criteria, including positive and negative criteria, as privileged teacher context~\citep{bablani2026rubrics}.
Our All-Rubric SD baseline adapts this approach to agentic RL by retaining GRPO and using rollout-derived rubrics.
\rised differs by combining distillation with prompt-group selection, while assigning positive rubrics to teacher conditioning and negative rubrics to rollout guidance. 

\section{FAQ}

\subsection{Why use a fixed rubric vocabulary?}

Our goal requires comparing rubric profiles across environments and training
steps. Dynamic generation of free-form rubric assigns semantically equivalent behaviours
different names and therefore requires repeated clustering or alignment before
profiles can be compared. In preliminary experiments, clustering free-form annotations yielded between 25 and 44 skills across checkpoints, making comparisons across steps and environments difficult. The verification of each step's new rubrics can also be difficult. A fixed vocabulary instead provides a stable
coordinate system for online comparison.

\subsection{Will the warm-up process in building vocabulary introduce additional time cost?}
The warm-up run, taking only $1/10$ steps of full training, introduces a one-time preprocessing cost that is amortized across subsequent training runs:

\begin{enumerate}
\item \textbf{Reuse across backbones.}
The warm-up policy provides failure trajectories used to discover negative rubrics. Once constructed, the resulting vocabulary is reused across training runs and model backbones, so a separate warm-up run is not required for each backbone.

\item \textbf{Reduced recurring overhead.}
A predefined vocabulary reduces online tagging to label selection. In contrast, free-form extraction would require repeated name alignment or clustering to reconcile semantically equivalent labels and preserve comparability across rollouts and training steps.

\item \textbf{Offline use of larger models.}
Rubric consolidation and description generation are performed offline using a larger model, while online tagging uses a locally served judge. This avoids repeated large-model inference during training and reduces throughput overhead.

\end{enumerate}

\subsection{Why evaluate rubric similarity at the environment level?}
We use environment-level analysis as a computationally tractable diagnostic of whether rubric similarity tracks reward improvements during single-environment training. Directly estimating the marginal benefit of individual prompt groups would require more expensive controlled updates and evaluations. The aggregate analysis motivates the predefined and shared rubric representation; it does not directly estimate each group’s contribution. We assess the practical utility of prompt-group selection through the main experiment and the end-to-end component ablations.

\subsection{Does reward-conditioned tagging affect the interpretation of rubric similarity?}
The judge receives rollout rewards and uses outcome-dependent instructions to identify desired and undesired behaviours. The resulting rubric profiles therefore reflect both behavioural content and outcome composition. The gradient baseline also incorporates rewards through its objective. Our comparison evaluates how well these two reward-informed representations track observed environment-level reward improvements. The stronger correlations support rubric similarity as an empirical diagnostic in this setting, but do not establish that its association is independent of reward composition.

\subsection{Are negative rubrics injected during validation?}
No. Negative rubrics guide training rollouts only. All reported validation results use the original task prompts without rubric injection. The role-filter ablation in App.~\ref{app:distill-role} and main results in Sec.~\ref{sec:exp:main} evaluate these training choices under the same unguided validation protocol. Its results show that the benefits of training-time guidance persist when the guidance is removed at evaluation.

\subsection{Does every group with identical rewards receive positive-rubric supervision?}
\label{app:faq:no_positive}
Not necessarily. Positive-rubric self-distillation can provide additional supervision for groups with identical rewards when positive rubrics are available within the group. Groups without positive rubrics lack this privileged behavioral context, and the distillation term has zero loss and gradients when teacher and student parameters and inputs coincide. However, negative rubrics can still provide a complementary mechanism by guiding subsequent rollout generation. 

\subsection{Why not evaluate larger policy backbones?}
Agent RL is already computationally expensive at the 3--4B scale.
For \texttt{Qwen2.5-3B-Instruct}, each run comprises $1000$ training
steps on $32$ H100 GPUs. Based on the measured average step times,
even GRPO-64 requires approximately $1.8$ days per run
($155.5$\,s/step), while \rised requires approximately $2.9$ days
($252.9$\,s/step). Validation every $10$ steps further increases the total runtime. Repeating training across baselines, ablations, and seeds
incurs substantial computational cost. We therefore prioritize
controlled comparisons across provided two backbones.

\end{document}